\documentclass{article}

    \PassOptionsToPackage{numbers, compress}{natbib}
\usepackage{neurips_data_2024}

\usepackage[utf8]{inputenc} % allow utf-8 input
\usepackage[T1]{fontenc}    % use 8-bit T1 fonts
\usepackage{hyperref}       % hyperlinks
\usepackage{url}            % simple URL typesetting
\usepackage{booktabs}       % professional-quality tables
\usepackage{amsfonts}       % blackboard math symbols
\usepackage{nicefrac}       % compact symbols for 1/2, etc.
\usepackage{microtype}      % microtypography
\usepackage[dvipsnames]{xcolor}         % colors
\usepackage{enumitem}
\usepackage{arydshln}
\usepackage{graphicx}
\usepackage{multirow}
\usepackage{adjustbox}
\usepackage{makecell}
\usepackage{subfigure}
\usepackage{wrapfig}
\usepackage{graphicx}
\usepackage{caption}  
\usepackage{amsmath}
\title{MobileAIBench: Benchmarking LLMs and LMMs for On-Device Use Cases}

\author{%
  Rithesh Murthy$^{1,\ast}$, Liangwei Yang$^{1,\ast}$, Juntao Tan$^{1,\ast}$, Tulika Manoj Awalgaonkar$^{1,}$\thanks{Equal Contribution.} \\
  \textbf{Yilun Zhou}$^{1}$, \textbf{Shelby Heinecke}$^{1}$, \textbf{Sachin Desai}$^{2}$,
  \textbf{Jason Wu}$^{1}$,
  \textbf{Ran Xu}$^{1}$, \textbf{Sarah Tan}$^{1}$, \\
  \textbf{Jianguo Zhang}$^{1}$,
    \textbf{Zhiwei Liu}$^{1}$,
  \textbf{Shirley Kokane}$^{1}$,
  \textbf{Zuxin Liu}$^{1}$,
  \textbf{Ming Zhu}$^{1}$, \\
  \textbf{Huan Wang}$^{1}$,
  \textbf{Caiming Xiong}$^{1}$,
  \textbf{Silvio Savarese}$^{1}$\\
  $^1$Salesforce AI Research \\
  $^2$Salesforce Mobile Platform \\
  \vspace{-1em}
  }

\begin{document}

\maketitle

\begin{abstract}
% Large Language Models (LLMs) and Large Multimodal Models (LMMs) have achieved remarkable performance across various tasks.
The deployment of Large Language Models (LLMs) and Large Multimodal Models (LMMs) on mobile devices has gained significant attention due to the benefits of enhanced privacy, stability, and personalization. However, the hardware constraints of mobile devices necessitate the use of models with fewer parameters and model compression techniques like quantization. Currently, there is limited understanding of quantization's impact on various task performances, including LLM tasks, LMM tasks, and, critically, trust and safety. There is a lack of adequate tools for systematically testing these models on mobile devices. To address these gaps, we introduce MobileAIBench, a comprehensive benchmarking framework for evaluating mobile-optimized LLMs and LMMs. MobileAIBench assesses models across different sizes, quantization levels, and tasks, measuring latency and resource consumption on real devices. Our two-part open-source framework includes a library for running evaluations on desktops and an iOS app for on-device latency and hardware utilization measurements. Our thorough analysis aims to accelerate mobile AI research and deployment by providing insights into the performance and feasibility of deploying LLMs and LMMs on mobile platforms.

\end{abstract}

\section{Introduction}
With billions of parameters trained on massive amounts of data, LLMs and LMMs have achieved remarkable breakthroughs in a wide range of applications from question answering~\cite{baek2023knowledge} to intelligent agents~\cite{wang2024survey,liu2024agentlite} and beyond ~\cite{achiam2023gpt,team2023gemini,nijkamp2022codegen}. Most recently, the pursuit of deploying LLMs and LMMs on mobile devices has garnered attention, and for good reason. There are several key benefits to deploying AI on mobile devices including offline access, enhanced privacy, and improved performance. It also provides cost efficiency by decreasing server and bandwidth usage, while enhancing user experience with faster and more interactive applications.

Given the extreme limitations of mobile hardware, deploying LLMs and LMMs on mobile devices is challenging. First, the model must have a relatively small number of parameters since parameters drive the number of computations, which consume memory, CPU, and GPU resources.
% There are a number of small open and closed source LLMs and LMMs emerging, such as Gemma-2B \cite{team2024gemma}, MobileLLM \cite{liu2024mobilellm}, and Gemini \cite{team2023gemini}. 
Second, even with a relatively small number of parameters, these models may not fit onto a mobile phone’s limited hardware. To further reduce resource footprint, quantization has emerged as a practical, heuristic approach to reduce precision of model weights with seemingly little penalty to performance.

While deploying LLMs and LMMs to mobile devices for real use cases appears to be feasible in the near future, there are knowledge and tooling gaps remaining. First, while quantization seems to be a practical way to reduce the resource footprint of small LLMs and LMMs, there is little to no  rigorous measurement and understanding of the effect of quantization on task performance, including LLM tasks, LMM tasks, and critically, trust and safety. Second, there is limited or no tooling available to systematically test quantized models across these tasks. Third, there is limited or no tooling available to test quantized models on a real mobile device across tasks. 

In this work, we aim to help accelerate mobile AI research and deployment by providing thorough benchmarking and analysis of open source mobile-optimized LLMs and LMMs. We restrict LLMs and LMMs in consideration to have at most 7B parameters, as we found 7B to be the upper limit of what a high-end phone’s hardware can manage (even after quantization). 
We measure and analyze current LLM and LMM task performance under different levels of quantization, from 16-bit quantization down to 3-bit quantization in some cases. We selected tasks that are most representative of real-world mobile use cases and considerations. In addition to task performance, we also benchmark our selected LLMs and LLMs on a real mobile device, an iPhone 14. We measure several key latency metrics such as time-to-first-token, and hardware utilization such as CPU usage and RAM usage. 

Our results are collected using MobileAIBench, our new two-part framework for evaluating LLMs and LMMs for mobile deployment. The first part of the framework is an open source library, for use on desktops or servers, to evaluate model's performance on a specially selected set of widely known benchmarks. Using this part of the framework, users are able to test their quantized models across benchmarks as they desire. The second part of the MobileAIBench framework is an open source iOS app. With our iOS app, users are able to measure latency and mobile hardware utilization such as RAM and CPU of quantized LLMs and LMMs. Our main contributions are summarized as follows:
\begin{itemize}
\item We are the first to provide a thorough benchmarking and analysis of open source LLMs and LMMs across varying levels of quantization and various tasks. Our evaluations are generated and reproducible using our newly developed framework, MobileAIBench.
\item 
% The release of MobileAIBench, our two-part framework for evaluating LLMs and LMMs for mobile deployment. 
% MobileAIBench is the first open source framework to enable task-specific LLM and LMM testing on-device. MobileAIBench will empower researchers to evaluate their small LLMs and LMMs and empower practitioners to assess viability of available LLMs and LMMs for mobile deployment.
MobileAIBench is the first open-source framework for on-device task-specific LLM and LMM testing, enabling researchers to evaluate small models and practitioners to assess model viability for mobile deployment.
\item We conduct extensive experiments to evaluate LLMs/LMMs over a wide range of tasks, providing insightful findings regarding the impact of quantization and real-mobile deployment.
\end{itemize}

\section{Related Work}
%\subsection{LLM Benchmarks}
Many benchmarks have been developed to evaluate LLMs and LMMs from different perspectives. For example, MMLU~\cite{hendrycks2020measuring} provides a large number of tasks to extensively test world knowledge and problem solving ability. AlpacaEval~\cite{dubois2024alpacafarm} and MT-Bench~\cite{zheng2024judging} provide open-ended question answering evaluation tasks without explicit answers, and employ GPT-4~\cite{achiam2023gpt} as the success rate judge. KoLA~\cite{yu2023kola} uses Wikipedia and the continuously collected emerging corpora data to provides knowledge-oriented LLM assessment tasks. TruthfulQA~\cite{lin2021truthfulqa}, TrustLLM~\cite{sun2024trustllm}, Safetybench~\cite{zhang2023safetybench} measure LLMs' trust and safety levels. To assist the development of agents, benchmarks~\cite{zhou2023instruction,chen2024benchmarking,zhou2023instruction} have also been developed to evaluate LLM's instruction-following ability. FOFO~\cite{xia2024fofo} contains diverse data formats, and is able to test the format-following ability of current LLMs. MME~\cite{fu2024mme} provides comprehensive multimodal evaluation benchmars over $14$ subtasks.
VisIT-Bench \cite{bitton2023visit} provides an instruction-following vision-language datasets to test LMMs' real-world use-case.
MMVP \cite{tong2024eyes} provides $9$ basic visual patterns that LMMs easily give incorrect answers and hallucinated explanations.
However, these benchmarks do not consider the quantized versions of models, nor determine the impact of deployment constraints on model performance. MobileAIBench fills this gap by focusing on deployment utilization on real mobile devices. Besides model performance on specific tasks, MobileAIBench emphasizes model quantization, inference speed, and required deployment resources.

%Different from previous benchmarks that test LLMs functionality ability, the proposed MobileAIBench focusing on LLM and LMM's deployment utilization on real mobile devices. Besides model performance on specific tasks, MobileAIBench emphasizes model quantization, inference speed and required deployment resources on mobile devices.

Several papers have discussed strategies and evaluations for developing mobile-ready models~\cite{zhang2024mobileenv}. \cite{jin2024comprehensive} compares different quantization methods and evaluates the performance of quantized LLMs. \cite{liu2024mobilellm} considered different architectures to develop the most performant mobile models.
MobileVLM~\cite{chu2023mobilevlm} designs vision language models for mobile devices.
The Octopus series~\cite{chen2024octopus} aim to empower the agentic ability on mobile device by training API tokens.
Recent small models such as Phi3 \cite{abdin2024phi} and Gemma \cite{team2024gemma} have, in their model cards, various evaluation results. However, none of these are as comprehensive as the set of data, models, and evaluation metrics proposed in MobileAIBench. In developing this standardized benchmark, we hope to make it easier for model developers to consider the various factors needed to develop mobile-ready models.

\section{MobileAIBench Framework}
MobileAIBench is a two-part evaluation framework for evaluating LLMs and LMMs for mobile deployment. The first part is a pipeline for use on desktops or servers, to evaluate model performance on a specially selected set of widely known benchmarks. The second part is an open source iOS app to measure latency and mobile hardware utilization.

% In this section, we give more details on the currently available components of both parts of our MobileAIBench framework. 

\begin{figure}[htbp]
    \centering
    % linewidth
    % textwidth
    \includegraphics[width=1.0\textwidth]{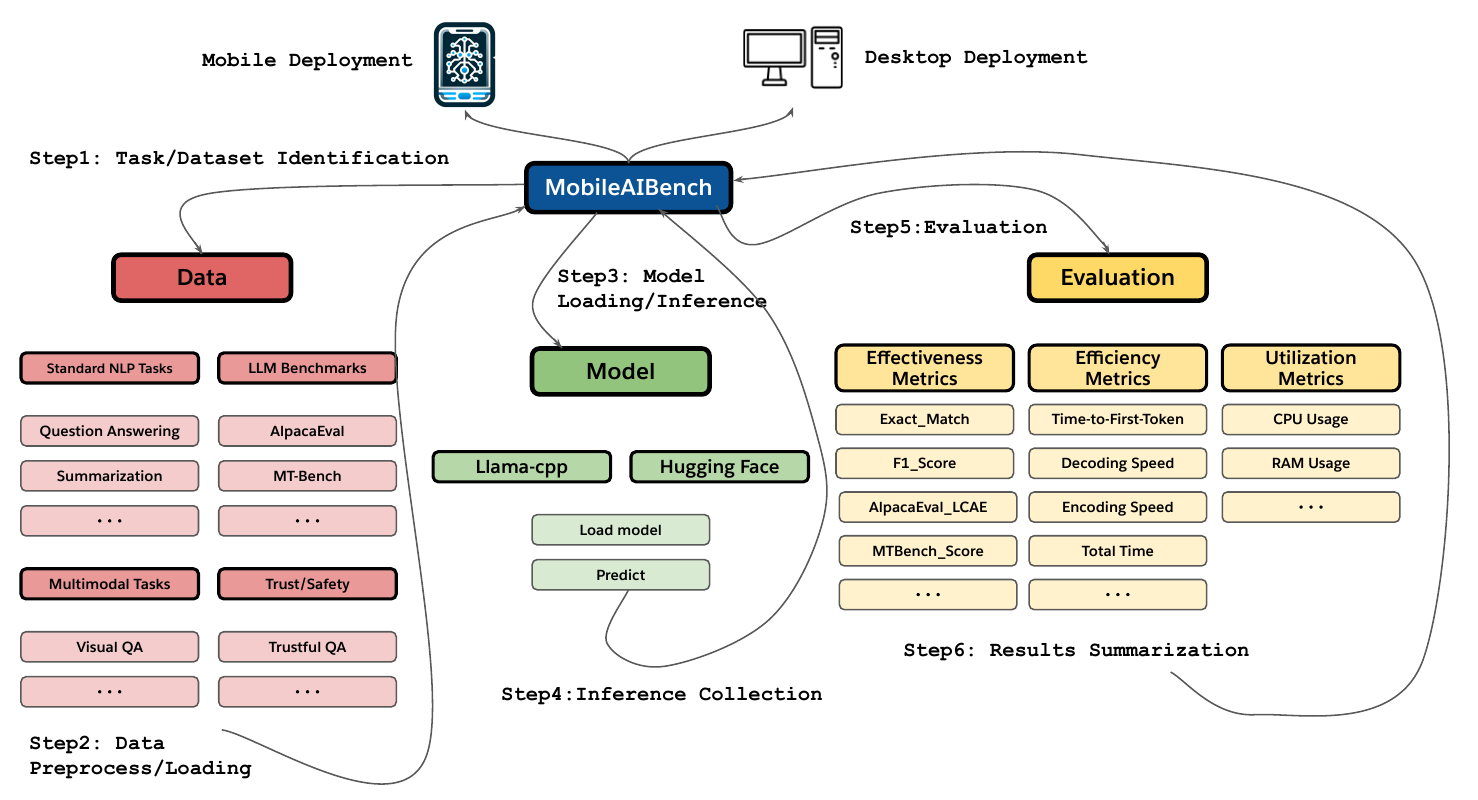}
    \captionsetup{justification=centering}
    \caption{MobileAIBench Architecture}
    \label{fig:pipeline}
\end{figure}

\subsection{Tasks and Datasets}\label{sec:tasks}
This section outlines the tasks and datasets used to benchmark LLMs \& LMMs across various domains including Natural Language Processing (NLP), multimodal, and trust \& safety tasks.
\subsubsection{Standard NLP Tasks}
Standard NLP tasks encompass various benchmarks designed to evaluate different capabilities of LLMs. Question answering tests an LLM's ability to comprehend context and respond accurately, for which we use the \textbf{Databricks-dolly} \cite{DatabricksBlog2023DollyV2} and \textbf{HotpotQA} \cite{yang2018hotpotqa} datasets. Summarization tasks involve condensing large amounts of information into shorter forms while retaining essential ideas. We assess LLM performance in summarization using the \textbf{CNN/Daily Mail} \cite{hermann2015teaching, nallapati2016abstractive} and \textbf{XSum} \cite{Narayan2018DontGM} datasets. Text-to-SQL tasks evaluate an LLM's proficiency in crafting SQL queries based on natural language questions. For this purpose, we employ the \textbf{Sql-Create-Context} \cite{b-mc2_2023_sql-create-context} dataset. Additionally, we include popular benchmarks such as the \textbf{Massive Multitask Language Understanding (MMLU)} \cite{hendrycks2021measuring} to evaluate the LLM's accuracy in multitask performance and \textbf{Grade School Math (GSM8K)} \cite{cobbe2021gsm8k} for assessing LLM's ability to solve mathematical problems. To further evaluate LLM performance, we utilize benchmarks such as \textbf{AlpacaEval}, an automatic evaluation method that quantifies the Win-Rate by measuring the proportion of instances where the model's output is preferred over the reference output \cite{alpaca_eval}, and \textbf{MT-Bench}, a collection of complex multi-turn open-ended questions used to evaluate chat assistants, with GPT-4 serving as the judge \cite{zheng2023judging}. For Question Answering, Summarization, Text-to-SQL, GSM8K and MMLU tasks, we randomly select 1,000 samples from each relevant dataset. For AlpacaEval and MT-Bench, we use the standard test sets.

\subsubsection{Multimodal Tasks}
% Multimodal tasks require LMMs to process and generate responses across different data types such as text, images, and audio. This is critical for developing AI systems that handle complex user requirements on mobile devices. We focus on \textbf{Visual Question Answering (VQA)} \cite{antol2015vqa}, selecting five datasets that cover a wide range of contexts. The datasets are roughly categorized into two types. (1) Direct Answer VQA includes datasets requiring the LMM to answer visual questions with a single word or phrase. The test datasets include \textbf{VQA-v2} \cite{goyal2017making}, \textbf{VizWiz} \cite{gurari2018vizwiz}, \textbf{GQA} \cite{hudson2019gqa}, and \textbf{TextVQA} \cite{singh2019towards}. (2) Multiple-Choice VQA involves selecting the correct answer from multiple choices, as seen in the \textbf{ScienceQA} \cite{lu2022learn} dataset. Similar to the standard NLP tasks, for each dataset, we randomly select 1000 samples to evaluate the LMMs' performance.
Multimodal tasks require LMMs to process different data types such as text, images, and audio. This is critical for developing AI systems that handle complex user requirements on mobile devices. We focus on \textbf{Visual Question Answering (VQA)} \cite{antol2015vqa}, selecting five datasets that cover a wide range of contexts. Among them, \textbf{VQA-v2} \cite{goyal2017making}, \textbf{VizWiz} \cite{gurari2018vizwiz}, \textbf{GQA} \cite{hudson2019gqa}, and \textbf{TextVQA} \cite{singh2019towards} require the LMM to directly answer visual questions with a single word or phrase. \textbf{ScienceQA} \cite{lu2022learn} dataset requires selecting the correct answer from multiple choices. Similar to the standard NLP tasks, for each dataset, we randomly select 1000 samples to evaluate the LMMs' performance.

\subsubsection{Trust and Safety}
To assess the societal impact of LLMs, we include a suite of trust and safety evaluations, focusing on six categories: truthfulness, safety, robustness, fairness, privacy, and ethics, following \citet{sun2024trustllm}. For truthfulness, we use the \textbf{TruthfulQA} \cite{lin2021truthfulqa} (TruthQA) dataset to assess the ability to select the correct answer from common misconceptions. For safety, the \textbf{Do-Not-Answer} \cite{wang2023not} (DNA) dataset measures if LLMs can refuse illegal, unethical, or otherwise undesirable requests. Robustness is evaluated using \textbf{Adversarial Instruction} \cite{sun2024trustllm} (Adv-Inst), which tests models' robustness to prompt perturbations like typos or irrelevant links. Fairness is measured using the \textbf{BBQ} dataset \cite{parrish2021bbq}, assessing the tendency to fall for common stereotypes related to gender, age, race, etc. Privacy is tested with hand-crafted prompts based on the \textbf{Enron email dataset} \cite{shetty2004enron} (Priv-Lk), examining if models can decline requests for personal information. For ethics, we use the \textbf{Social Chemistry 101} (SC-101) dataset \cite{forbes2020social} to evaluate moral acceptability judgments for different situations.

\subsection{Part 1: Evaluation Pipeline for Desktop and Cloud}

The MobileAIBench pipeline, as shown in Figure \ref{fig:pipeline}, encompasses three main stages: Data, Model, and Evaluation. In the Data stage, the task and relevant datasets are identified, and the evaluation dataset is created through preprocessing and prompt hydration before being fed into the Model stage. In the Model stage, the model is initialized and predictions are made on the evaluation data, which are subsequently assessed in the Evaluation stage. Various task-specific and generic metrics are supported at the Evaluation stage to gauge the performance of the LLMs. Additionally, MobileAIBench serves as a versatile tool for other researchers and developers to construct their own benchmarking frameworks, thanks to its plug-and-play design, which allows for the easy addition of new tasks and metrics and the creation of custom leaderboards.

% \textbf{Performance Metrics} are evaluated on Desktop/Cloud for faster inference to obtain the performance on different tasks. The performance results evaluated are the same on mobile devices. These metrics are task-oriented, and evaluate model's effectiveness on different tasks. Same as previous benchmarks, these metrics aim to show model's performance on varied tasks.

\textbf{Performance Metrics} are evaluated on desktop for faster inference, with results consistent on mobile devices. These task-oriented metrics assess model effectiveness across various tasks.

\subsection{Part 2: Mobile App}\label{sec:app}

\begin{wrapfigure}{r}{0.3\textwidth}
\vspace{-10pt}
% \setlength{\abovecaptionskip}{-8pt}
    % \centering
    \includegraphics[width=\linewidth]{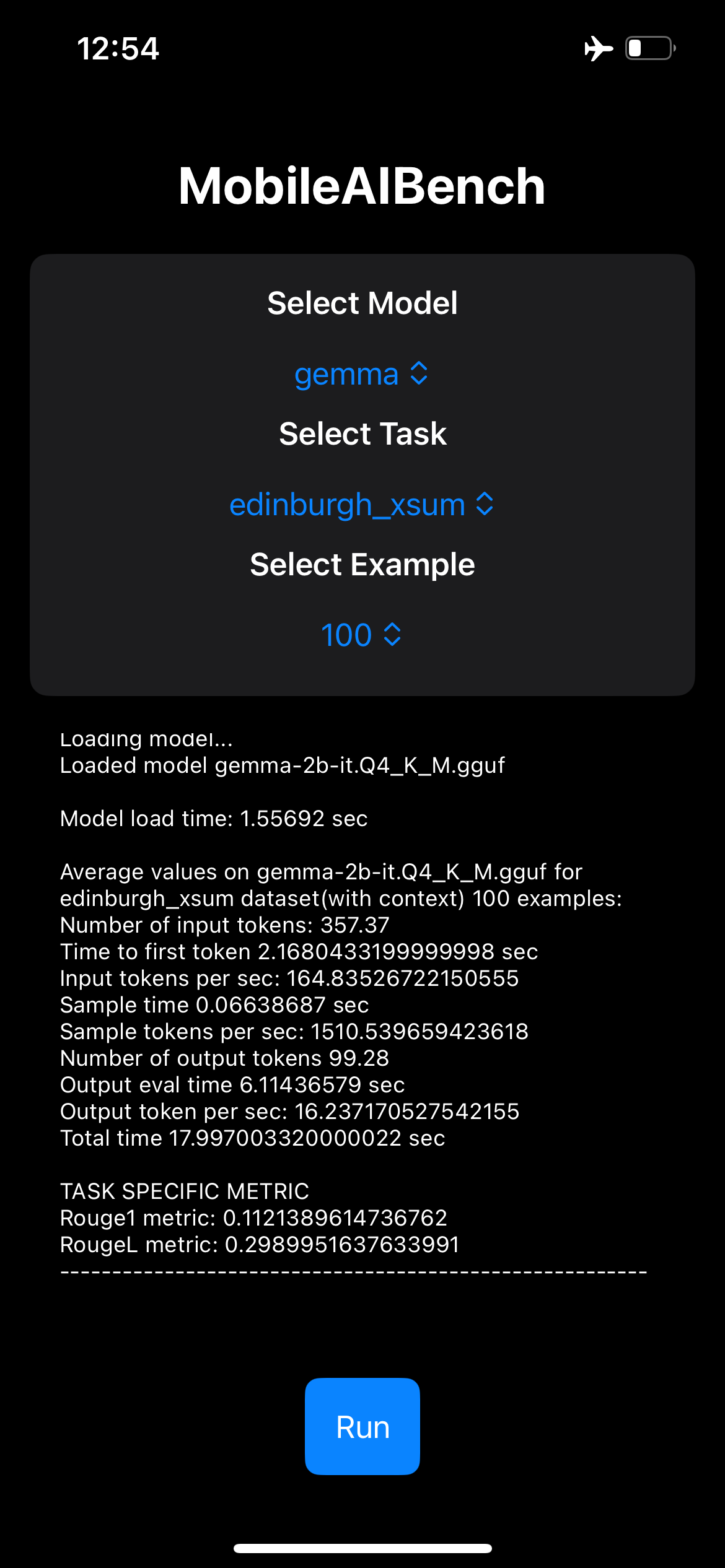}
    \captionsetup{justification=centering}
    \caption{\\ MobileAIBench iOS app}
    \label{fig:APP}
    \vspace{-38pt}
\end{wrapfigure}

The mobile app component of MobileAIBench is designed to extend the evaluation capabilities to actual mobile devices as shown in Figure~\ref{fig:APP}. This allows for a more accurate assessment of LLM and LMM performance in real-world scenarios. By utilizing the app, we can measure critical efficiency and utilization metrics directly on the device, providing insights into how these models will perform when deployed on end-user mobile hardware. This comprehensive evaluation ensures that the models are not only effective but also efficient and practical for mobile deployment.

\textbf{Mobile Efficiency \& Utilization Metrics} are evaluated on real mobile devices using the iOS app to test the deployment feasibility of LLMs and LMMs. \textbf{Efficiency} metrics include Time-to-First-Token (TTFT, s), Input Token Per Second (ITPS, t/s), Output Evaluation Time (OET, t), and Total Time (s). These metrics are model-oriented and measure the efficiency of the models when running on mobile devices. The efficiency is influenced by multiple factors, including model structure, quantization level, and prompt template. \textbf{Utilization} metrics measure the resource consumption when running models on real mobile devices. These device-oriented metrics consist of CPU, RAM usage and Battery Drain Rate (BDR, \%).

\subsection{Model Libraries Supported}
MobileAIBench is designed to seamlessly test different models on mobile devices. It integrates two inference libraries to accommodate a wide range of LLMs \& LMMs: (1) Huggingface\footnote{https://huggingface.co/}, which allows users to test any models available on pre-defined tasks by simply changing the model name. Inference with Huggingface provides the performance of the original pre-trained models. (2) Llama.cpp\footnote{https://github.com/ggerganov/llama.cpp}, which allows users to test models on real mobile devices. The Llama.cpp inference method supports quantization to reduce model sizes, facilitating deployment on mobile devices.

\section{Benchmarks and Analysis}
% In this section, we present the comprehensive experiments conducted using MobileAIBench. We discuss the experimental setup, models chosen for evaluation, the different quantization levels examined, the benchmark results obtained, and our subsequent analysis.
For Effectiveness and Trust \& Safety experiments, we conducted evaluations on a desktop to assess the impact of various quantization levels. These results are consistent when tested on mobile devices. To evaluate Efficiency and Resource Utilization, we tested the models' performance on an iPhone 14 using our iOS app. The process for selecting evaluation models is detailed in Section~\ref{sec:model_selection}.

\subsection{Effectiveness Evaluation}\label{sec:effectiveness}
In this section, we examine the effectiveness of various models with different quantization levels.

% \begin{table}[h]\centering
% \caption{Effectiveness of LLMs Across Various Benchmarks and Trust \& Safety Metrics}
% \begin{adjustbox}{width=1.0\linewidth}
% \begin{tabular}{cccccccccccccc}
% \toprule
% \multirow{2}{*}{Quantization} & \multirow{2}{*}{Model} & \multirow{2}{*}{\makecell{Model Size \\ Category}} & \multirow{2}{*}{Disk Usage} & \multirow{2}{*}{AlpacaEval} & \multirow{2}{*}{MT-Bench} & \multirow{2}{*}{MMLU} & \multicolumn{1}{c}{\makecell{Math \\ Reasoning }} & \multicolumn{6}{c}{Trust \& Safety} \\
% \cmidrule(lr){8-9}  \cmidrule(lr){9-14} 
%  & & & & & & & GSM8K & TQA & BBQ & SC-101 & Ad-I & DNA & Pr-L \\
% \midrule

% \begin{table}[h]\centering
% \caption{Effectiveness of LLMs Across Various Benchmarks and Trust \& Safety Metrics}
% \begin{adjustbox}{width=1.0\linewidth}
% \begin{tabular}{
%     c c c c c c c @{\hskip 5pt} c @{\hskip 5pt} c @{\hskip 5pt} c @{\hskip 5pt} c @{\hskip 5pt} c @{\hskip 5pt} c @{\hskip 5pt} c
% }
% \toprule
% \multirow{2}{*}{Quantization} & \multirow{2}{*}{Model} & \multirow{2}{*}{\makecell{Model Size \\ Category}} & \multirow{2}{*}{Disk Usage} & \multirow{2}{*}{AlpacaEval} & \multirow{2}{*}{MT-Bench} & \multirow{2}{*}{MMLU} & \multicolumn{1}{c}{\makecell{Math \\ Reasoning }} & \multicolumn{6}{c}{Trust \& Safety} \\
% \cmidrule(lr){8-8}  \cmidrule(lr){9-14} 
%  & & & & & & & GSM8K & TQA & BBQ & SC-101 & Ad-I & DNA & Pr-L \\
% \midrule

% across multiple tasks, using task-specific metrics for evaluation.
\subsubsection{Standard NLP Tasks}

\begin{table}[h]\centering
\caption{Effectiveness of LLMs across standard NLP tasks.}
\begin{adjustbox}{width=1.0\linewidth}
\begin{tabular}{cccccccccccccc}
\toprule
\multirow{3}[3]{*}{Quantization} & \multirow{3}[3]{*}{Model} & \multirow{3}[3]{*}{\makecell{Model Size \\ Category}} & \multirow{3}[3]{*}{Disk Usage} &
\multicolumn{4}{c}{Question Answering} & \multicolumn{2}{c}{Text-to-SQL} & \multicolumn{4}{c}{Summarization}\\
\cmidrule(lr){5-8} \cmidrule(lr){9-10} \cmidrule(lr){11-14}
 & & & & \multicolumn{2}{c}{Databricks} & \multicolumn{2}{c}{HotpotQA} & \multicolumn{2}{c}{sql-create-context} & \multicolumn{2}{c}{CNN} & \multicolumn{2}{c}{XSum} \\
 \cmidrule(lr){5-6} \cmidrule(lr){7-8} \cmidrule(lr){9-10} \cmidrule(lr){11-12} \cmidrule(lr){13-14}
 & & & & EM & F1 & EM & F1 & SP & LS & R1 & RL & R1 & RL \\
 \midrule
 \multirow{7}{*}{16bit} 
& Llama 2 7B    & $>$ 6B    & 13 GB & 0.034 & 0.443 & 0.071 & \underline{0.210} & 0.492 & 0.845 & 0.322 & 0.204 & 0.174 & 0.118 \\
& Mistral 7B    & $>$ 6B    & 14 GB & \underline{0.043} & \textbf{0.498} & \textbf{0.137} & \textbf{0.267} & 0.485 & 0.770 & 0.328 & 0.204 & 0.170 & 0.114 \\
& Gemma 7B      & $>$ 6B    & 16 GB & 0.026 & \underline{0.479} & 0.000 & 0.098 & \textbf{0.546} & \textbf{0.856} & 0.336 & \underline{0.218} & 0.187 & \underline{0.127} \\
& Phi 2 3B      & 1B - 6B   & 5.2 GB & \textbf{0.046} & 0.472 & \underline{0.096} & 0.197 & 0.489 & \underline{0.852} & \underline{0.352} & 0.216 & \textbf{0.220} & \textbf{0.145} \\
& Gemma 2B      & 1B - 6B   & 4.7 GB & 0.025 & 0.401 & 0.001 & 0.046 & \underline{0.519} & 0.849 & \textbf{0.368} & \textbf{0.241} & \underline{0.214} & \textbf{0.145} \\
& Zephyr 3B     & 1B - 6B   & 5.3 GB & 0.032 & 0.441 & 0.030 & 0.109 & 0.457 & 0.787 & 0.325 & 0.203 & 0.169 & 0.112 \\
& TinyLlama 1B  & 1B - 6B   & 2.1 GB & 0.003 & 0.329 & 0.000 & 0.076 & 0.389 & 0.719 & 0.320 & 0.202 & 0.170 & 0.113 \\
\midrule

\multirow{7}{*}{8bit} 
& Llama 2 7B    & $>$ 6B    & 6.7 GB & 0.038 & 0.444 & 0.071 & \underline{0.208} & 0.491 & 0.845 & 0.323 & 0.205 & 0.173 & 0.117 \\
& Mistral 7B    & $>$ 6B    & 7.2 GB & \underline{0.045} & \textbf{0.507} & \underline{0.089} & \textbf{0.209} & \underline{0.519} & \textbf{0.854} & 0.352 & \underline{0.228} & 0.183 & 0.122 \\
& Gemma 7B      & $>$ 6B    & 8.5 GB & 0.026 & \underline{0.476} & 0.001 & 0.097 & \textbf{0.544} & \textbf{0.854} & 0.337 & 0.219 & 0.188 & 0.128 \\
& Phi 2 3B      & 1B - 6B   & 2.8 GB & \textbf{0.047} & 0.472 & \textbf{0.099} & 0.202 & 0.493 & \underline{0.852} & \underline{0.353} & 0.216 & \textbf{0.217} & \textbf{0.144} \\
& Gemma 2B      & 1B - 6B   & 2.5 GB & 0.024 & 0.398 & 0.003 & 0.049 & 0.518 & 0.849 & \textbf{0.368} & \textbf{0.239} & \underline{0.213} & \underline{0.143} \\
& Zephyr 3B     & 1B - 6B   & 2.8 GB & 0.032 & 0.440 & 0.030 & 0.108 & 0.449 & 0.782 & 0.324 & 0.203 & 0.170 & 0.113 \\
& TinyLlama 1B  & 1B - 6B   & 1.1 GB & 0.002 & 0.322 & 0.000 & 0.069 & 0.359 & 0.710 & 0.315 & 0.198 & 0.166 & 0.110 \\
\midrule
 \multirow{7}{*}{4bit} 
& Llama 2 7B    & $>$ 6B    & 3.6 GB & 0.033 & 0.445 & 0.055 & \underline{0.198} & 0.490 & 0.841 & 0.322 & 0.204 & 0.173 & 0.116 \\
& Mistral 7B    & $>$ 6B    & 3.9 GB & \textbf{0.042} & \textbf{0.499} & \textbf{0.150} & \textbf{0.273} & 0.474 & 0.774 & 0.325 & 0.202 & 0.170 & 0.113 \\
& Gemma 7B      & $>$ 6B    & 4.7 GB & 0.021 & \underline{0.470} & 0.000 & 0.096 & \textbf{0.549} & \textbf{0.859} & 0.336 & \underline{0.218} & 0.189 & 0.128 \\
& Phi 2 3B      & 1B - 6B   & 1.5 GB & \underline{0.039} & 0.466 & \underline{0.082} & 0.190 & 0.465 & 0.837 & \underline{0.342} & 0.209 & \textbf{0.214} & \underline{0.141} \\
& Gemma 2B      & 1B - 6B   & 1.5 GB & 0.017 & 0.384 & 0.001 & 0.041 & \underline{0.510} & \underline{0.844} & \textbf{0.366} & \textbf{0.237} & \underline{0.213} & \textbf{0.144} \\
& Zephyr 3B     & 1B - 6B   & 1.5 GB & 0.035 & 0.445 & 0.027 & 0.106 & 0.459 & 0.789 & 0.330 & 0.206 & 0.170 & 0.112 \\
& TinyLlama 1B  & 1B - 6B   & 0.6 GB & 0.002 & 0.348 & 0.000 & 0.070 & 0.407 & 0.735 & 0.323 & 0.202 & 0.176 & 0.116 \\
\bottomrule
\end{tabular}
\end{adjustbox}
\end{table}

We use several evaluation metrics to assess the performance of the models across different tasks. For question answering tasks, we employ Exact Match (EM) and F1 Score (F1). In the context of Text-to-SQL tasks, we utilize the SQL Parser (SP) and Levenshtein Score (LS). For summarization tasks, we measure performance using Rouge-1 (R1) and Rouge-L (RL). Additionally, we use Win-Rate for AlpacaEval, Score for MT Bench, and Accuracy for both MMLU and GSM8K. Detailed explanations on the implementation of these metrics, along with additional information, are provided in Appendix \ref{sec:eval_metrics}. The performance of various models across different quantization levels is presented in Table 1 and Table 2. In these tables, the highest score for each quantization category is indicated in bold, while the second-best score is underlined. Figure 5 illustrates the violin plots depicting the performance changes when models are quantized from 16-bit to 8-bit. Specifically, Figure 5(a) shows the distribution of performance changes for each model across different tasks, including standard NLP tasks and trust \& safety tasks. Figure 5(b) presents the distribution of performance changes for each task when the underlying model shifts from 16-bit to 8-bit.

\textbf{Observation and Analysis:} The results indicate that no single model consistently outperforms all others across every task. However, on average, large LLMs (> 6B parameters) exhibit superior performance compared to medium LLMs (1B-6B parameters). While quantization does introduce some performance changes, these changes are not significant in most cases. This finding enhances our confidence in deploying quantized models on mobile devices without substantial performance degradation. In figure 5(a), a narrow distribution and smaller range of performance changes indicate a model's robustness to quantization, meaning these models are relatively less sensitive to the quantization process. In contrast, a more spread-out distribution suggests greater sensitivity to quantization, as there is more deviation in performance. The graphs clearly show that different models respond differently to quantization, with some being more sensitive than others. Figure 5(b) demonstrates the robustness of various tasks to the quantization of underlying models, highlighting the importance of considering both model size and sensitivity to quantization when deploying models on edge devices to ensure optimal performance across diverse tasks.

\subsubsection{Multi-modality Tasks}
Table~\ref{tab:LMM_main} shows each model's performance on the selected VQA datasets. The evaluation prompt and evaluation metrics for each dataset can be seen in Section ~\ref{sec:prompt_template} and Section ~\ref{sec:eval_metrics}.

\textbf{Observation and Analysis:} Under the original precision (16-bit), no single model outperforms the others across all datasets. On average, Llava-v1.5-7B and BakLLava outperform the others, indicating that larger models have advantages for visual-language understanding. We specifically note that although Moondream2 has only around $1.7$B parameters, its performance is highly comparable to the 7B models. It only falls short on SQA, which is the only dataset providing related context besides the image and questions for the models to effectively answer the questions. This may indicate that even the strongest smaller models lack context understanding ability.

\begin{figure}[htbp]
    \centering
    \begin{minipage}[b]{0.42\linewidth}
        \centering
        \includegraphics[width=\linewidth]{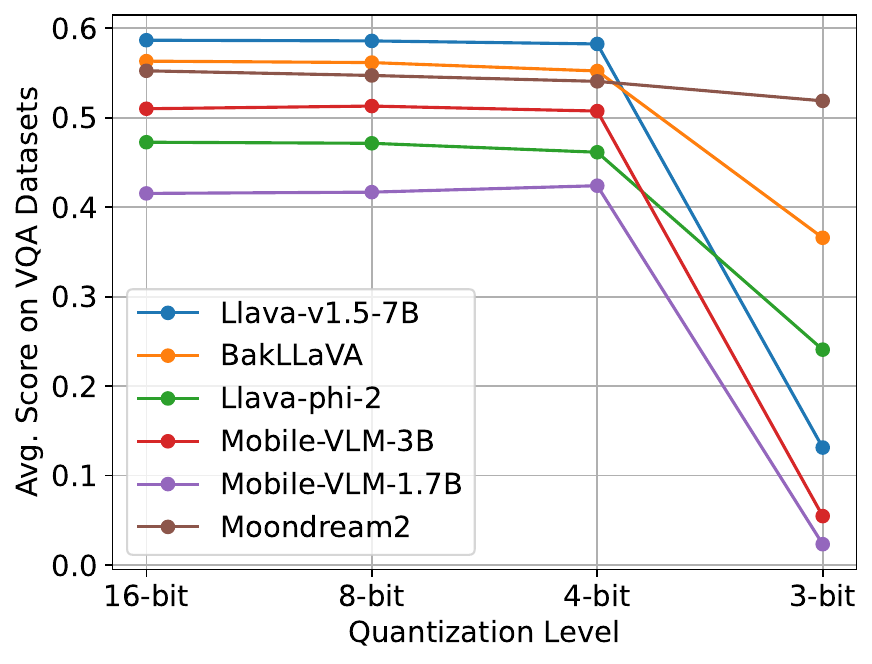}
        \caption{Performance change of LMMs under different quantization.}
        \label{fig:LMM_quant}
    \end{minipage}
    % \hspace{-5pt}
    \hfill
    \begin{minipage}[b]{0.48\linewidth}
        \centering
        \includegraphics[width=\linewidth]{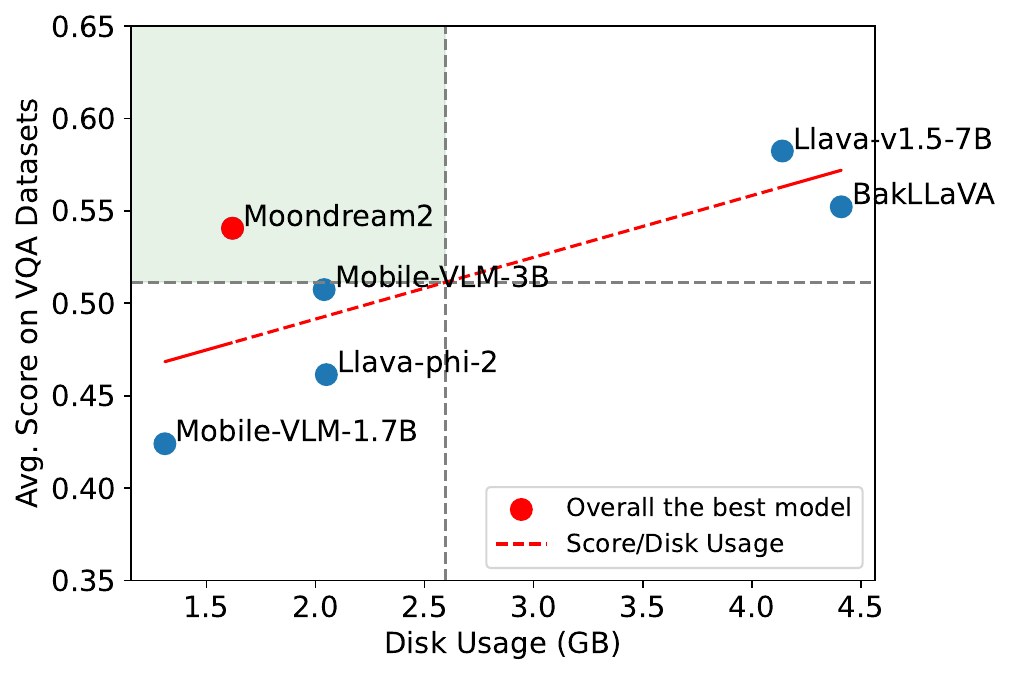}
        \caption{Trade-off between accuracy and disk usage under 4-bit quantization.}
        \label{fig:LMM_du}
    \end{minipage}
\end{figure}

We observe that the models' accuracy remains consistent across different quantization levels until 3-bit quantization, where most models experience a significant performance drop, as shown in Figure \ref{fig:LMM_quant}. Moreover, Moondream2 is surprisingly robust to quantization, even at the 3-bit level. This indicates that the effect of quantization on LMMs can vary significantly. Given the importance of quantization for on-device AI, evaluating different models' robustness to quantization is crucial and should be a focus of further study in the AI community.

Disk usage is also an important aspect when deploying models on mobile devices. Therefore, we conducted further evaluation on the trade-off between accuracy and disk usage. As shown in Figure \ref{fig:LMM_du}, a linear trend indicates that the performance of LLMs generally increases with disk usage, which is expected as disk usage is highly related to the number of model parameters. The models in the top-left quadrant are considered the best overall regarding both accuracy and size.

% Models above the trend line can be seen as those with more satisfactory overall performance.

% Considering it small size, it could be one of the most suitable model to be deployed on devices. The e

% 1: main. no model consistantly good on all dataset. 7b good, moondream, newer.
% 2: quantization 16, 8, 4 stable, 3 drop. Moondream2 is extamly robust to quantization, even at 3-bit, which means... Second best bakllaba. The evaluation of different models' robustness to quantization is very important but not fully awared by the community. Fully study the reason of this in expected in the future
% 3: 4-bit is the most feasible to be fit into mobile devices. Trade-offs.

% By looking into the performance of original 16-bit models, Llava-v1.5-7b 

% LLM main results Table \ref{tab:LMM_main}. Performance change with Quantization in Figure \ref{fig:LMM_quant}. Performance vs disk usage \ref{fig:LMM_du}

% ScienceQA less sensitive to quantization (simply generation, quantize deeper loss more generation ability compared to visual language understanding)

% Add analysis: which model loss more on quantization (bar), word/phrase based answer. Multiple choice answer

% TTFT vs. Quant trend

% Appendix: Latency analysis on CPU, evaluation metrics, evaluation prompts

\begin{table}[htbp]
\centering
% \caption{Comparison of Multimodal Models with Different Quantization Methods}
\caption{Effectiveness of LMMs across various VQA datasets.}
\label{tab:LMM_main}
\begin{adjustbox}{width=1.0\linewidth}
\begin{tabular}{ccccccccc|c}
\toprule
Quantization & Model     & Model Size & Disk Usage & VQA-v2 & GQA & VisWiz & TextVQA & SQA & Avg.\\ \midrule
\multirow{6}{*}{16bit}           & Llava-v1.5-7B      & $>$ 6B & 13.13 GB &  0.760 & \underline{0.596}  & \textbf{0.545}  & \underline{0.416}  & 0.616 & \textbf{0.587}\\ 
                                 & BakLLaVA           &       $>$ 6B                  & 14.07 GB                 &  \underline{0.770}                 &  \textbf{0.602}                 & 0.385                  & 0.407                  & \underline{0.652}          & \underline{0.563}    \\ 
                                 & Llava-phi-2        & 1B - 6B & 5.74 GB                 &  0.658               &  0.484                 & 0.269                 & 0.298                 & \textbf{0.654}    & 0.473          \\ 
                                 & Mobile-VLM-3B      &         1B - 6B                 & 5.63 GB                 &  0.713                 & 0.552                 &  0.448                & 0.337                  & 0.500        & 0.510      \\ 
                                 & Mobile-VLM-1.7B    & 1B - 6B & 3.12 GB                 & 0.622                  & 0.509                  & 0.315                  & 0.234                  &  0.397        & 0.415      \\ 
                                 & Moondream2         &         1B - 6B                 & 3.49 GB                  & \textbf{0.781}                  & 0.590                  &  \underline{0.470}                 & \textbf{0.441}                  &  0.480             & 0.552     \\ \midrule
\multirow{6}{*}{8bit}    & Llava-v1.5-7B      & $>$ 6B & 7.25 GB  &  0.759 & \textbf{0.600}  & \textbf{0.545}  & \underline{0.419}  & 0.606  & \textbf{0.586}\\ 
                                 & BakLLaVA           &            $>$ 6B             & 7.75 GB                  & \underline{0.770}                  &  \underline{0.597}                 & 0.384                 & 0.402                  & \underline{0.655}       & \underline{0.562}       \\ 
                                 & Llava-phi-2        & 1B - 6B & 3.31 GB                 & 0.660                  &  0.480                 & 0.260                  & 0.296                 & \textbf{0.661}       & 0.471         \\ 
                                 & Mobile-VLM-3B      &         1B - 6B                 &  3.27 GB                 &  0.714                 & 0.566                  &  \underline{0.460}                 &  0.335                & 0.490        & 0.513       \\ 
                                 & Mobile-VLM-1.7B    & 1B - 6B & 1.93 GB                  &  0.619                 &  0.515                 & 0.308                  &  0.241                 & 0.401         & 0.417      \\ 
                                 & Moondream2         &         1B - 6B                 &  2.26 GB                 &  \textbf{0.777}                 & 0.596                  &  0.457                 & \textbf{0.430}                  & 0.476          & 0.547      \\ \midrule
\multirow{6}{*}{4bit} & Llava-v1.5-7B      & $>$ 6B & 4.14 GB  & 0.750  & \underline{0.590}  & \textbf{0.536} & \textbf{0.414} & \underline{0.622}  & \textbf{0.583}\\ 
                                 & BakLLaVA           &    $>$ 6B                     & 4.41 GB                  & \underline{0.767}                  & \textbf{0.605}                  &  0.438                & 0.403                  & 0.548             & \underline{0.552}     \\ 
                                 & Llava-phi-2        & 1B - 6B & 2.05 GB                  & 0.658                  & 0.475                  & 0.247                  & 0.281                  & \textbf{0.646}        &   0.461   \\ 
                                 & Mobile-VLM-3B      &    1B - 6B                      & 2.04 GB                  &  0.705                 & 0.548                  & \underline{0.459}                  & 0.329                 & 0.496           & 0.507     \\ 
                                 & Mobile-VLM-1.7B    & 1B - 6B &  1.31 GB                 &  0.607                 & 0.500                &  0.375                 &  0.232                 & 0.406           & 0.424    \\ 
                                 & Moondream2         &        1B - 6B                  &  1.62 GB                 & \textbf{0.780}                  &  0.587                 & 0.446                  & \underline{0.411}                  &  0.479           & 0.541\\ \midrule
\multirow{6}{*}{3bit} & Llava-v1.5-7B      & $>$ 6B & 3.33 GB  & 0.148  & 0.050  & 0.016  & 0.083  & 0.360  & 0.131\\ 
                                 & BakLLaVA           &          $>$ 6B               &  3.53 GB               &  \underline{0.532}                 & \underline{0.450}                  & \underline{0.035}                 & \underline{0.223}                  & \textbf{0.589}         & \underline{0.366}     \\
                                 & Llava-phi-2        & 1B - 6B & 1.73 GB                 & 0.396                & 0.229                 & 0.011                 & 0.096                 & \underline{0.472}     & 0.144       \\ 
                                 & Mobile-VLM-3B      &     1B - 6B                     & 1.71 GB                & 0.146                & 0.031                  & 0.012                & 0.057                 & 0.028          & 0.055     \\ 
                                 & Mobile-VLM-1.7B    & 1B - 6B & 1.15 GB                 &  0.076                &  0.000                 &  0.008                 & 0.030                & 0.003        & 0.023      \\ 
                                 & Moondream2         &    1B - 6B                     &  1.46 GB                 &  \textbf{0.754}                 & \textbf{0.565}                  & \textbf{0.486}                 & \textbf{0.381}                  & 0.408       & \textbf{0.519}       \\ 
\bottomrule
\end{tabular}
\end{adjustbox}
\end{table}

% \begin{figure}[t]
% \centering
% \begin{adjustbox}{width=0.85\linewidth}
% \mbox{
%     \subfigure[]{
%         \includegraphics[width=0.48\textwidth]{figures/mm_vqa-v2.pdf}}
%     \subfigure[]{
%         \includegraphics[width=0.48\textwidth]{figures/mm_sqa.pdf}}
% }
% \end{adjustbox}
% \captionsetup{justification=centering}
% \caption{Performance change of LMMs on a) VQA-V2 and b) SQA under different quantization.}
% \label{fig:LMM_quant}
% \end{figure}

% \begin{figure}[t]
% \centering
% \begin{adjustbox}{width=0.85\linewidth}
% \mbox{
%     \subfigure[]{
%         \includegraphics[width=0.48\textwidth]{figures/mm_score_du_direct.pdf}}
%     \subfigure[]{
%         \includegraphics[width=0.48\textwidth]{figures/mm_score_du_mul.pdf}}
% }
% \end{adjustbox}
% \captionsetup{justification=centering}
% \caption{Performance of different LMMs under 4-bit quantization on a) Direct answer VQA, and b) Multiple-choice VQA. The models in the top-left quadrant are considered the best overall considering the trade-off between accuracy and disk usage.}
% \label{fig:LMM_du}
% \end{figure}

% \begin{figure}[t]
%     \centering
    
%     \begin{minipage}[b]{0.48\linewidth}
%         \centering
%         \includegraphics[width=\linewidth]{figures/vqa-v2.png}
%         \caption{Performance vs. Precision Levels on VQA-V2 Dataset.}
%     \end{minipage}
%     \hfill
%     \begin{minipage}[b]{0.48\linewidth}
%         \centering
%         \includegraphics[width=\linewidth]{figures/sqa.png}
%         \caption{Performance vs. Precision Levels on ScienceQA Dataset.}
%     \end{minipage}
% \end{figure}

\subsection{Trust \& Safety Evaluation}

\begin{table}[h]\centering
\caption{Effectiveness of LLMs across various NLP benchmarks and Trust \& Safety datasets.}
\begin{adjustbox}{width=1.0\linewidth}
% \fontsize{18}{18}\selectfont % Increase font size to 12pt with 15pt leading

\begin{tabular}{
    c c c c c c @{\hskip 5pt} c @{\hskip 5pt} c @{\hskip 10pt} c @{\hskip 10pt} c @{\hskip 10pt} c @{\hskip 10pt} c @{\hskip 10pt} c @{\hskip 10pt} c
}
\toprule
\multirow{2}{*}{Quantization} & \multirow{2}{*}{Model} & \multirow{2}{*}{\makecell{Model Size \\ Category}} & \multirow{2}{*}{Disk Usage} & \multicolumn{1}{c}{\makecell{Instruction \\ Task}} & \multicolumn{1}{c}{\makecell{Multi-turn \\ Chat}} & \multicolumn{1}{c}{\makecell{Multi-task \\ Eval}} & \multicolumn{1}{c}{\makecell{Mathematical \\ Reasoning}} & \multicolumn{6}{c}{Trust \& Safety} \\
\cmidrule(lr){5-5} \cmidrule(lr){6-6} \cmidrule(lr){7-7} \cmidrule(lr){8-8} \cmidrule(lr){9-14} 
 & & & & AlpacaEval & MT-Bench & MMLU & GSM8K & TruthQA &  BBQ  & SC-101 & Adv-Inst & DNA & Priv-Lk \\
\midrule
% \fontsize{20}{15}\selectfont % Increase font size to 12pt with 15pt leading
\multirow{7}{*}{16bit} 
& Llama 2 7B    & $>$ 6B    & 13 GB & 4.918 &	\underline{6.259} & 0.418 & 0.271 & 0.272 & 0.341 & 0.626 & \textbf{0.943} & \textbf{0.855} & \textbf{1.000} \\
& Mistral 7B    & $>$ 6B    & 14 GB & \textbf{12.185} &	\textbf{7.484} & \textbf{0.536} & \textbf{0.509} & \textbf{0.512} & \textbf{0.783} & 0.706 & \underline{0.925} & 0.589 & \underline{0.993} \\
& Gemma 7B      & $>$ 6B    & 16 GB & 1.116 &	4.959 & 0.464 & 0.400 & 0.294 & 0.544 & \textbf{0.736} & 0.892 & 0.788 & \textbf{1.000} \\
& Phi 2 3B      & 1B - 6B   & 5.2 GB & 2.932 &	5.318 & \underline{0.479} & 0.190 & \underline{0.370} & 0.588 & 0.676 & 0.849 & 0.277 & 0.020 \\
& Gemma 2B      & 1B - 6B   & 4.7 GB & 6.459 &	5.187 & 0.352 & 0.135 & 0.187 & 0.310 & 0.594 & 0.916 & \underline{0.830} & \textbf{1.000} \\
& Zephyr 3B     & 1B - 6B   & 5.3 GB & \underline{8.261} &	6.009 & 0.391 & \underline{0.491} & 0.300 & \underline{0.621} & \underline{0.716} & 0.860 & 0.608 & \textbf{1.000} \\
& TinyLlama 1B  & 1B - 6B   & 2.1 GB & 1.529 &	3.962 & 0.168 & 0.110 & 0.153 & 0.236 & 0.628 & 0.879 & 0.366 & 0.530 \\

\midrule
\multirow{7}{*}{8bit} 
& Llama 2 7B    & $>$ 6B    & 6.7 GB & 5.184 &	5.182 & 0.423 & 0.277 & 0.272 & 0.336 & 0.636 & \textbf{0.942} & \textbf{0.854} & \textbf{1.000} \\
& Mistral 7B    & $>$ 6B    & 7.2 GB & \textbf{8.508} &	\textbf{6.428} & \underline{0.474} & \underline{0.466} & \underline{0.343} & 0.497 & \textbf{0.744} & \underline{0.912} & 0.390 & \underline{0.987} \\
& Gemma 7B      & $>$ 6B    & 8.5 GB & 1.339 &	4.693 & 0.469 & 0.220 & 0.294 & 0.531 & \underline{0.740} & 0.888 & 0.790 & \textbf{1.000} \\
& Phi 2 3B      & 1B - 6B   & 2.8 GB & 2.972 &	5.740 & \textbf{0.482} & 0.229 & \textbf{0.367} & \underline{0.580} & 0.674 & 0.847 & 0.274 & 0.27 \\
& Gemma 2B      & 1B - 6B   & 2.5 GB & 5.713 &	5.328 & 0.352 & 0.146 & 0.192 & 0.307 & 0.592 & 0.910 & \underline{0.831} & \textbf{1.000} \\
& Zephyr 3B     & 1B - 6B   & 2.8 GB & \underline{7.682} &	\underline{6.256} & 0.386 & \textbf{0.499} & 0.290 & \textbf{0.620} & 0.714 & 0.864 & 0.608 & \textbf{1.000} \\
& TinyLlama 1B  & 1B - 6B   & 1.1 GB & 1.514 &	3.562 & 0.188 & 0.160 & 0.158 & 0.256 & 0.612 & 0.846 & 0.389 & 0.367 \\

\midrule
\multirow{7}{*}{4bit} 
& Llama 2 7B    & $>$ 6B    & 3.6 GB & 4.172 &	\underline{6.146} & 0.403 & 0.263 & 0.291 & 0.368 & 0.594 & \textbf{0.925} & \textbf{0.861} & \textbf{1.000} \\
& Mistral 7B    & $>$ 6B    & 3.9 GB & \textbf{13.269} &	\textbf{7.600} & \textbf{0.521} & \underline{0.464} & \textbf{0.499} & \textbf{0.776} & 0.700 & \underline{0.915} & 0.588 & \underline{0.993} \\
& Gemma 7B      & $>$ 6B    & 4.7 GB & 1.713 &	4.728 & 0.465 & 0.310 & 0.291 & 0.488 & \textbf{0.730} & 0.883 & 0.802 & \textbf{1.000} \\
& Phi 2 3B      & 1B - 6B   & 1.5 GB & 3.583 &	5.750 & \underline{0.477} & 0.224 & \underline{0.371} & \underline{0.613} & 0.702 & 0.817 & 0.309 & 0.087 \\
& Gemma 2B      & 1B - 6B   & 1.5 GB & 4.835 &	4.946 & 0.346 & 0.117 & 0.181 & 0.289 & 0.588 & 0.914 & \underline{0.817} & \textbf{1.000} \\
& Zephyr 3B     & 1B - 6B   & 1.5 GB & \underline{6.980} &	5.959 & 0.371 & \textbf{0.496} & 0.289 & 0.596 & \underline{0.712} & 0.842 & 0.609 & 0.993 \\
& TinyLlama 1B  & 1B - 6B   & 0.6 GB & 1.320 &	3.534 & 0.167 & 0.140 & 0.135 & 0.252 & 0.636 & 0.859 & 0.363 & 0.080 \\
\bottomrule
\end{tabular}
\end{adjustbox}
\end{table}

\begin{figure}[htbp]
\centering
\begin{adjustbox}{width=1.0\linewidth}
\mbox{
    \subfigure[]{
        \includegraphics[width=0.50\textwidth]{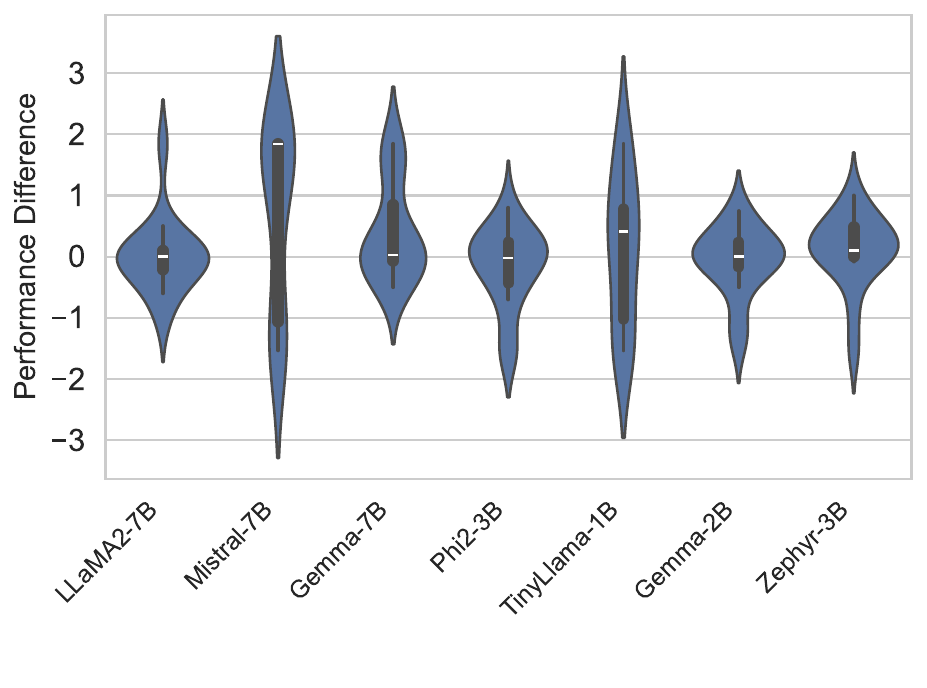}}
    \subfigure[]{
        \includegraphics[width=0.50\textwidth]{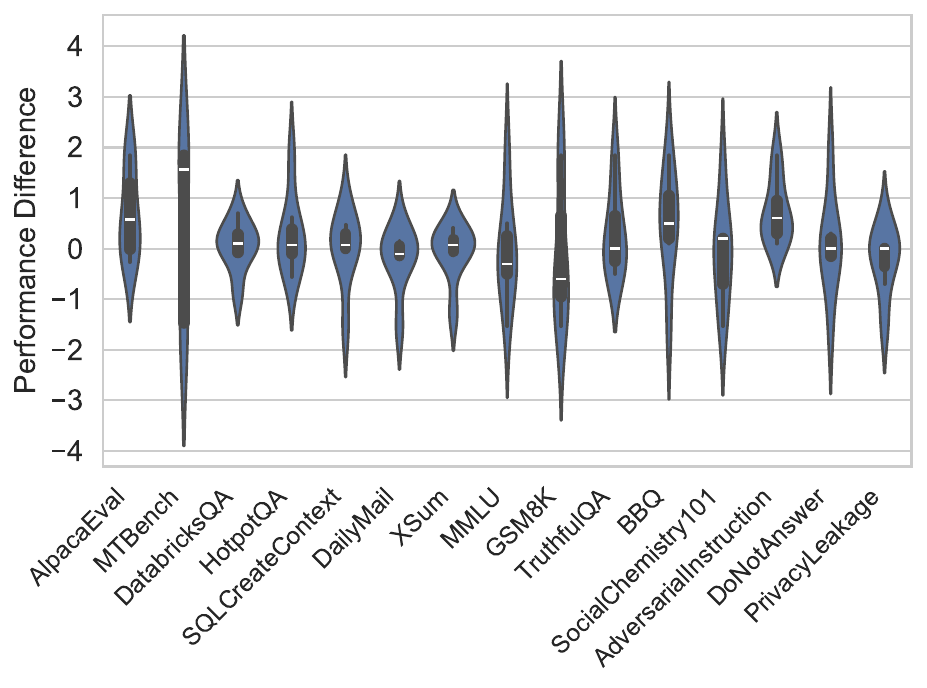}}
}
\end{adjustbox}
\captionsetup{justification=centering}
\caption{Distribution of performance changes: (a) per LLM, (b) per task, when transitioning from 16-bit to 8-bit quantization.}
\end{figure}

We evaluate the performance of various models on trust and safety-related tasks. Ensuring robust performance in these areas is critical, as NLP models are increasingly deployed in sensitive and high-stakes environments. We evaluate the models on various datasets, as discussed in Section 3.1.3. Accuracy is used as the primary effectiveness metric, and we utilize GPT-4o in a llm-as-a-judge framework to determine the accuracy across each of these datasets. The results are shown in Table 3. 

\textbf{Observation and Analysis:} As with standard NLP tasks, no single model consistently excels across all trust and safety tasks. Larger LLMs (>6B parameters) generally perform better than medium-sized LLMs (1B-6B parameters). While quantization does affect performance, the impact is minimal, indicating that quantized models can be effectively used in trust and safety applications without significant performance degradation.

\subsection{Efficiency and Utilization Evaluation}

We evaluate current LLMs and LMMs on real mobile device using the iOS app provided within MobileAIBench to test the efficiency and utilization. All experiments are evaluated on the same iPhone 14 device to guarantee the comparability. All models are quantized to 4bit and only those under 3B are deployable. We conduct the experiments with $3$ LLMs (Phi2 3B, Gemma 2B, TinyLlama 1B) on $4$ NLP datasets and $1$ LMM (Llava-Phi-2) on $2$ LMM datasets.

\begin{table}[h]
\centering
\caption{Efficiency \& Utilization of LLMs across NLP tasks. Metrics explanation in Section~\ref{sec:app}}
% Metrics explanation: Time to First Token (TTFT), Input Token per Second (TIPS), Output Evaluation Time (OET), Output Token per Second (OTPS), Battery Drain Rate (BDR)}
\label{tab:efficient_llm}
\begin{adjustbox}{width=1.0\linewidth}
\begin{tabular}{cccccccccc}
\toprule
\multirow{2}[3]{*}{Dataset} & \multirow{2}[3]{*}{Model} &
\multicolumn{5}{c}{Efficiency} & \multicolumn{3}{c}{Utilization} \\
\cmidrule(lr){3-7} \cmidrule(lr){8-10}
& & TTFT(s) & ITPS(t/s) & OET(s) & OTPS(t/s) & Total Time(t) & CPU(\%)  & RAM(GiB) & BDR(\%) \\

 \midrule
 \multirow{3}{*}{HotpotQA} 
& Phi2 3B    & 2.32 & 94.04 & \textbf{1.78} & 13.21 & \textbf{5.02} & \textbf{63.89} & 4.33 & 9.33\\
& Gemma 2B    & 2.86 & 133.35 & 3.52 & 13.65 & 11.62 & 85.3 & 4.25 & 10.22\\
& TinyLlama 1B      & \textbf{1.60} & \textbf{277.24} & 3.12 & \textbf{28.14} & 6.30 & 71.58 & \textbf{3.34} & \textbf{5.37}\\
\midrule
\multirow{3}{*}{Databricks-dolly} 
& Phi2 3B    & 2.01 & 88.39 & 3.35 & 12.74 & 7.01 & \textbf{61.28} & 4.17 & 11.76\\
& Gemma 2B    & 1.93 & 168.49 & \textbf{2.83} & 16.31 & 9.25 & 87.66 & 4.24 & 18.51\\
& TinyLlama 1B      & \textbf{1.08} & \textbf{345.32} & 2.88 & \textbf{31.24} & \textbf{5.45} & 70.47 & \textbf{3.34} & \textbf{10.00}\\
\midrule
\multirow{3}{*}{Sql-create-context} 
& Phi2 3B    & 0.73 & 96.53 & \textbf{1.72} & 13.84 & \textbf{3.36}& \textbf{76.67} & 4.37 & 9.09\\
& Gemma 2B    & 0.75 & 163.01 & 2.14 & 16.77 & 6.56 & 99.17 & 4.45 & 21.6\\
& TinyLlama 1B      & \textbf{0.39} & \textbf{349.05} & 2.08 & \textbf{32.89} & 3.61 & 80.31 & \textbf{3.37} & \textbf{7.50}\\
\midrule
\multirow{3}{*}{XSum} 
& Phi2 3B    & 2.73 & 73.40 & 8.41 & 11.66 & 15.13 & \textbf{68.35} & 4.57 & 18.66\\
& Gemma 2B    & 2.30 & 154.88 & 6.37 & 15.56 & 18.99 & 94.00 & 4.44 & 36.17\\
& TinyLlama 1B      & \textbf{1.29} & \textbf{321.70} &\textbf{4.10} & \textbf{30.13} & \textbf{7.45} & 70.88 & \textbf{3.43} & \textbf{15.00}\\
% \midrule 
% \hline
% VQA-V2 & Llava-Phi-2 & & & & & & 47.68 & 4.69 \\
% ScienceQA & Llava-Phi-2 & & & & & & & \\
\bottomrule
\end{tabular}
\end{adjustbox}
\end{table}

\textbf{Observation and Analysis:} Experiment results of LLMs are shown in Table~\ref{tab:efficient_llm}. We can have the following observations. (1) Smaller models have lower TTFT and higher ITPS/OTPS. It indicates smaller models have faster encoding/decoding speed as the computation required for processing each token is decreased. 
(2) The shortest OET and Total Time may not be achieved by the smallest model. For example, on HotpotQA dataset, the lowest OET and Total Time are achieved by Phi2 model. It owes to the input/output token length that is directly related with the model. Though Phi2 has the slowest encoding/decoding speed, it can achieve the fastest OET and Total time by giving more concise responses.
(3) The on-device memory consumption is intense and directly relates to model sizes. For the iPhone 14 with a total RAM of $6$ GiB, even running the 4-bit quantized TinyLlama model (1B) on-device takes more than $50\%$ of the overall memory, leaving limited space for other APPs. The memory consumption increases with larger models.
(4) The CPU utilization is naturally different with different models. On all the $4$ datasets, Phi2 takes the lowest and Gemma takes the highest CPU utilization, respectively. It is a surprise finding that the on-device CPU utilization is not related to model sizes, which reals the necessity of on-device testing for mobile deployment.
(5) Battery Drain Rate (BDR) increases with both model size and the number of output tokens generated. Larger models and longer outputs consume more battery power, highlighting the need for optimization in model design and output length to improve energy efficiency.

\begin{wraptable}{r}{0.618\textwidth}
\vspace{-10pt}
\centering
\caption{Efficiency \& Utilization of LMMs.}
\label{tab:efficiency_lmm}
\begin{adjustbox}{width=1.0\linewidth}
\begin{tabular}{ccccccc}
\toprule
\multirow{2}[3]{*}{Dataset} & \multirow{2}[3]{*}{Model} & \multirow{2}[3]{*}{Samples} &
\multicolumn{2}{c}{Efficiency} & \multicolumn{2}{c}{Utilization} \\
\cmidrule(lr){4-5} \cmidrule(lr){6-7}
& & & TTFT(s) & ITPS(t/s) & CPU(\%)  & RAM(GiB) \\
\midrule
\multirow{3}{*}{VQA-v2} 
& \multirow{3}{*}{Llava-Phi-2} & 10 & \textbf{66.47} & \textbf{1.24}& \textbf{51.06} & \textbf{4.63}\\
& & 25 & 213.48 & 0.37 & 51.29 & 4.66\\
& & 50 & 350.06 & 0.22 &  57.62 & 4.67\\
\midrule
\multirow{3}{*}{ScienceQA} 
& \multirow{3}{*}{Llava-Phi-2} & 10 & \textbf{81.00} & \textbf{4.65} & 90.39 & \textbf{4.56} \\
& & 25 & 223.61 & 1.55 & 78.46 & 4.63\\
& & 50 & 508.66 & 0.68 & \textbf{77.52} & 4.65\\
\bottomrule
\end{tabular}
\end{adjustbox}
\vspace{-10pt}
\end{wraptable}

The results for LMMs are presented in Table \ref{tab:efficiency_lmm}. Since the number of output tokens is only 1 for both datasets, we do not have OET and OPTS efficiency results, and the TTFT is equivalent to the Total Time. As observed, multimodal tasks are significantly more computation-intensive compared to NLP tasks. The average TTFT exceeds 60 seconds for both tasks, and the ITPS is lower than 5. This indicates that current LMMs may not yet be suitable for mobile deployment. However, this could improve with newer mobile devices with enhanced computational capabilities. Additionally, we notice a decrease in efficiency with an increasing number of samples, likely due to the elevated temperature resulting from processing more samples. In Section \ref{sec:LMM_cpu}, we conduct a comprehensive latency analysis of the LMMs on Intel CPUs to further compare their latency with each other. The RAM usage keeps nearly constant after loading the model, which also matches the observation of NLP tasks in Section~\ref{sec:utilization}. 
% Also, for different datasets, the CPU utilization of the same model also differs, which shows the impact of prompt length for different datasets.

\section{Discussion}\label{sec:discussion}

MobileAIBench offers a comprehensive evaluation of LLMs and LMMs in terms of task effectiveness, on-device efficiency, and utilization. It covers a wide range of text and multimodal tasks, assessing the impact of various quantization levels on current models. Additionally, trust and safety evaluations are included to ensure the reliability and security of LLM/LMM applications on mobile devices. As a benchmarking platform for mobile deployment, MobileAIBench aims to enhance performance while minimizing any potential negative social impact.

Extensive experiments with MobileAIBench reveal several interesting findings. Quantization is an effective way to decrease model size while keeping the performance. Different models/tasks have varied sensitivity to quantization levels. Current LLMs and LMMs require significant resources in terms of CPU and RAM usage when deployed on mobile devices, even with the 1B model. More compact and optimized LLMs and LMMs are needed for mobile deployment.

While MobileAIBench is a significant advancement, we acknowledge its limitations. It excludes other model compression methods, like model pruning, as they are not mature enough for deployment. Additionally, due to space constraints and the wide variety of mobile devices, we have not conducted a comprehensive comparison across different hardware. Support for different species of mobile devices is still under development and will open to public upon release. 
% Although MobileAIBench stands as a significant advancement, we also acknowledge its limitations. It leaves other model compression methods behind such as the model pruning as those methods are still not mature enough for real deployment. Due to the space limitation and the wide range of mobile devices, we do not have a thorough comparing on different hardwares. Support for different species of mobile devices is still under development and will open to public upon release. 

\bibliographystyle{plainnat}
\bibliography{references}

\begin{thebibliography}{51}
\providecommand{\natexlab}[1]{#1}
\providecommand{\url}[1]{\texttt{#1}}
\expandafter\ifx\csname urlstyle\endcsname\relax
  \providecommand{\doi}[1]{doi: #1}\else
  \providecommand{\doi}{doi: \begingroup \urlstyle{rm}\Url}\fi

\bibitem[Abdin et~al.(2024)Abdin, Jacobs, Awan, Aneja, Awadallah, Awadalla, Bach, Bahree, Bakhtiari, Behl, et~al.]{abdin2024phi}
Marah Abdin, Sam~Ade Jacobs, Ammar~Ahmad Awan, Jyoti Aneja, Ahmed Awadallah, Hany Awadalla, Nguyen Bach, Amit Bahree, Arash Bakhtiari, Harkirat Behl, et~al.
\newblock Phi-3 technical report: A highly capable language model locally on your phone.
\newblock \emph{arXiv preprint arXiv:2404.14219}, 2024.

\bibitem[Achiam et~al.(2023)Achiam, Adler, Agarwal, Ahmad, Akkaya, Aleman, Almeida, Altenschmidt, Altman, Anadkat, et~al.]{achiam2023gpt}
Josh Achiam, Steven Adler, Sandhini Agarwal, Lama Ahmad, Ilge Akkaya, Florencia~Leoni Aleman, Diogo Almeida, Janko Altenschmidt, Sam Altman, Shyamal Anadkat, et~al.
\newblock Gpt-4 technical report.
\newblock \emph{arXiv preprint arXiv:2303.08774}, 2023.

\bibitem[Antol et~al.(2015)Antol, Agrawal, Lu, Mitchell, Batra, Zitnick, and Parikh]{antol2015vqa}
Stanislaw Antol, Aishwarya Agrawal, Jiasen Lu, Margaret Mitchell, Dhruv Batra, C~Lawrence Zitnick, and Devi Parikh.
\newblock Vqa: Visual question answering.
\newblock In \emph{Proceedings of the IEEE international conference on computer vision}, pages 2425--2433, 2015.

\bibitem[b~mc2(2023)]{b-mc2_2023_sql-create-context}
b~mc2.
\newblock sql-create-context dataset, 2023.
\newblock URL \url{https://huggingface.co/datasets/b-mc2/sql-create-context}.

\bibitem[Baek et~al.(2023)Baek, Aji, and Saffari]{baek2023knowledge}
Jinheon Baek, Alham~Fikri Aji, and Amir Saffari.
\newblock Knowledge-augmented language model prompting for zero-shot knowledge graph question answering.
\newblock \emph{arXiv preprint arXiv:2306.04136}, 2023.

\bibitem[Bitton et~al.(2023)Bitton, Bansal, Hessel, Shao, Zhu, Awadalla, Gardner, Taori, and Schimdt]{bitton2023visit}
Yonatan Bitton, Hritik Bansal, Jack Hessel, Rulin Shao, Wanrong Zhu, Anas Awadalla, Josh Gardner, Rohan Taori, and Ludwig Schimdt.
\newblock Visit-bench: A benchmark for vision-language instruction following inspired by real-world use.
\newblock \emph{arXiv preprint arXiv:2308.06595}, 2023.

\bibitem[Chen et~al.(2024{\natexlab{a}})Chen, Li, and Ma]{chen2024octopus}
Wei Chen, Zhiyuan Li, and Mingyuan Ma.
\newblock Octopus: On-device language model for function calling of software apis.
\newblock \emph{arXiv preprint arXiv:2404.01549}, 2024{\natexlab{a}}.

\bibitem[Chen et~al.(2024{\natexlab{b}})Chen, Xu, Wang, Liu, and Mao]{chen2024benchmarking}
Yihan Chen, Benfeng Xu, Quan Wang, Yi~Liu, and Zhendong Mao.
\newblock Benchmarking large language models on controllable generation under diversified instructions.
\newblock \emph{arXiv preprint arXiv:2401.00690}, 2024{\natexlab{b}}.

\bibitem[Chu et~al.(2023)Chu, Qiao, Lin, Xu, Yang, Hu, Wei, Zhang, Zhang, Wei, et~al.]{chu2023mobilevlm}
X~Chu, L~Qiao, X~Lin, S~Xu, Y~Yang, Y~Hu, F~Wei, X~Zhang, B~Zhang, X~Wei, et~al.
\newblock Mobilevlm: A fast, strong and open vision language assistant for mobile devices.
\newblock \emph{arXiv preprint arXiv:2312.16886}, 2023.

\bibitem[Cobbe et~al.(2021)Cobbe, Kosaraju, Bavarian, Chen, Jun, Kaiser, Plappert, Tworek, Hilton, Nakano, Hesse, and Schulman]{cobbe2021gsm8k}
Karl Cobbe, Vineet Kosaraju, Mohammad Bavarian, Mark Chen, Heewoo Jun, Lukasz Kaiser, Matthias Plappert, Jerry Tworek, Jacob Hilton, Reiichiro Nakano, Christopher Hesse, and John Schulman.
\newblock Training verifiers to solve math word problems.
\newblock \emph{arXiv preprint arXiv:2110.14168}, 2021.

\bibitem[Conover et~al.(2023)Conover, Hayes, Mathur, Xie, Wan, Shah, Ghodsi, Wendell, Zaharia, and Xin]{DatabricksBlog2023DollyV2}
Mike Conover, Matt Hayes, Ankit Mathur, Jianwei Xie, Jun Wan, Sam Shah, Ali Ghodsi, Patrick Wendell, Matei Zaharia, and Reynold Xin.
\newblock Free dolly: Introducing the world's first truly open instruction-tuned llm, 2023.
\newblock URL \url{https://www.databricks.com/blog/2023/04/12/dolly-first-open-commercially-viable-instruction-tuned-llm}.

\bibitem[Dubois et~al.(2024)Dubois, Li, Taori, Zhang, Gulrajani, Ba, Guestrin, Liang, and Hashimoto]{dubois2024alpacafarm}
Yann Dubois, Chen~Xuechen Li, Rohan Taori, Tianyi Zhang, Ishaan Gulrajani, Jimmy Ba, Carlos Guestrin, Percy~S Liang, and Tatsunori~B Hashimoto.
\newblock Alpacafarm: A simulation framework for methods that learn from human feedback.
\newblock \emph{Advances in Neural Information Processing Systems}, 36, 2024.

\bibitem[Forbes et~al.(2020)Forbes, Hwang, Shwartz, Sap, and Choi]{forbes2020social}
Maxwell Forbes, Jena~D Hwang, Vered Shwartz, Maarten Sap, and Yejin Choi.
\newblock Social chemistry 101: Learning to reason about social and moral norms.
\newblock \emph{arXiv preprint arXiv:2011.00620}, 2020.

\bibitem[Fu et~al.(2024)Fu, Chen, Shen, Qin, Zhang, Lin, Yang, Zheng, Li, Sun, Wu, and Ji]{fu2024mme}
Chaoyou Fu, Peixian Chen, Yunhang Shen, Yulei Qin, Mengdan Zhang, Xu~Lin, Jinrui Yang, Xiawu Zheng, Ke~Li, Xing Sun, Yunsheng Wu, and Rongrong Ji.
\newblock Mme: A comprehensive evaluation benchmark for multimodal large language models, 2024.

\bibitem[Goyal et~al.(2017)Goyal, Khot, Summers-Stay, Batra, and Parikh]{goyal2017making}
Yash Goyal, Tejas Khot, Douglas Summers-Stay, Dhruv Batra, and Devi Parikh.
\newblock Making the v in vqa matter: Elevating the role of image understanding in visual question answering.
\newblock In \emph{Proceedings of the IEEE conference on computer vision and pattern recognition}, pages 6904--6913, 2017.

\bibitem[Gurari et~al.(2018)Gurari, Li, Stangl, Guo, Lin, Grauman, Luo, and Bigham]{gurari2018vizwiz}
Danna Gurari, Qing Li, Abigale~J Stangl, Anhong Guo, Chi Lin, Kristen Grauman, Jiebo Luo, and Jeffrey~P Bigham.
\newblock Vizwiz grand challenge: Answering visual questions from blind people.
\newblock In \emph{Proceedings of the IEEE conference on computer vision and pattern recognition}, pages 3608--3617, 2018.

\bibitem[Hendrycks et~al.(2020)Hendrycks, Burns, Basart, Zou, Mazeika, Song, and Steinhardt]{hendrycks2020measuring}
Dan Hendrycks, Collin Burns, Steven Basart, Andy Zou, Mantas Mazeika, Dawn Song, and Jacob Steinhardt.
\newblock Measuring massive multitask language understanding.
\newblock \emph{arXiv preprint arXiv:2009.03300}, 2020.

\bibitem[Hendrycks et~al.(2021)Hendrycks, Burns, Basart, Zou, Mazeika, Song, and Steinhardt]{hendrycks2021measuring}
Dan Hendrycks, Collin Burns, Steven Basart, Andy Zou, Mantas Mazeika, Dawn Song, and Jacob Steinhardt.
\newblock Measuring massive multitask language understanding, 2021.

\bibitem[Hermann et~al.(2015)Hermann, Kocisky, Grefenstette, Espeholt, Kay, Suleyman, and Blunsom]{hermann2015teaching}
Karl~Moritz Hermann, Tomas Kocisky, Edward Grefenstette, Lasse Espeholt, Will Kay, Mustafa Suleyman, and Phil Blunsom.
\newblock Teaching machines to read and comprehend.
\newblock \emph{Advances in neural information processing systems}, 28, 2015.

\bibitem[Hudson and Manning(2019)]{hudson2019gqa}
Drew~A Hudson and Christopher~D Manning.
\newblock Gqa: A new dataset for real-world visual reasoning and compositional question answering.
\newblock In \emph{Proceedings of the IEEE/CVF conference on computer vision and pattern recognition}, pages 6700--6709, 2019.

\bibitem[Javaheripi et~al.(2023)Javaheripi, Bubeck, Abdin, Aneja, Bubeck, Mendes, Chen, Del~Giorno, Eldan, Gopi, et~al.]{javaheripi2023phi}
Mojan Javaheripi, S{\'e}bastien Bubeck, Marah Abdin, Jyoti Aneja, Sebastien Bubeck, Caio C{\'e}sar~Teodoro Mendes, Weizhu Chen, Allie Del~Giorno, Ronen Eldan, Sivakanth Gopi, et~al.
\newblock Phi-2: The surprising power of small language models.
\newblock \emph{Microsoft Research Blog}, 2023.

\bibitem[Jiang et~al.(2023)Jiang, Sablayrolles, Mensch, Bamford, Chaplot, Casas, Bressand, Lengyel, Lample, Saulnier, et~al.]{jiang2023mistral}
Albert~Q Jiang, Alexandre Sablayrolles, Arthur Mensch, Chris Bamford, Devendra~Singh Chaplot, Diego de~las Casas, Florian Bressand, Gianna Lengyel, Guillaume Lample, Lucile Saulnier, et~al.
\newblock Mistral 7b.
\newblock \emph{arXiv preprint arXiv:2310.06825}, 2023.

\bibitem[Jin et~al.(2024)Jin, Du, Huang, Liu, Luan, Wang, and Xiong]{jin2024comprehensive}
Renren Jin, Jiangcun Du, Wuwei Huang, Wei Liu, Jian Luan, Bin Wang, and Deyi Xiong.
\newblock A comprehensive evaluation of quantization strategies for large language models.
\newblock \emph{arXiv preprint arXiv:2402.16775}, 2024.

\bibitem[Li et~al.(2023)Li, Zhang, Dubois, Taori, Gulrajani, Guestrin, Liang, and Hashimoto]{alpaca_eval}
Xuechen Li, Tianyi Zhang, Yann Dubois, Rohan Taori, Ishaan Gulrajani, Carlos Guestrin, Percy Liang, and Tatsunori~B. Hashimoto.
\newblock Alpacaeval: An automatic evaluator of instruction-following models.
\newblock \url{https://github.com/tatsu-lab/alpaca\_eval}, 2023.

\bibitem[Lin et~al.(2021)Lin, Hilton, and Evans]{lin2021truthfulqa}
Stephanie Lin, Jacob Hilton, and Owain Evans.
\newblock Truthfulqa: Measuring how models mimic human falsehoods.
\newblock \emph{arXiv preprint arXiv:2109.07958}, 2021.

\bibitem[Liu et~al.(2024{\natexlab{a}})Liu, Zhao, Iandola, Lai, Tian, Fedorov, Xiong, Chang, Shi, Krishnamoorthi, et~al.]{liu2024mobilellm}
Zechun Liu, Changsheng Zhao, Forrest Iandola, Chen Lai, Yuandong Tian, Igor Fedorov, Yunyang Xiong, Ernie Chang, Yangyang Shi, Raghuraman Krishnamoorthi, et~al.
\newblock Mobilellm: Optimizing sub-billion parameter language models for on-device use cases.
\newblock \emph{arXiv preprint arXiv:2402.14905}, 2024{\natexlab{a}}.

\bibitem[Liu et~al.(2024{\natexlab{b}})Liu, Yao, Zhang, Yang, Liu, Tan, Choubey, Lan, Wu, Wang, et~al.]{liu2024agentlite}
Zhiwei Liu, Weiran Yao, Jianguo Zhang, Liangwei Yang, Zuxin Liu, Juntao Tan, Prafulla~K Choubey, Tian Lan, Jason Wu, Huan Wang, et~al.
\newblock Agentlite: A lightweight library for building and advancing task-oriented llm agent system.
\newblock \emph{arXiv preprint arXiv:2402.15538}, 2024{\natexlab{b}}.

\bibitem[Lu et~al.(2022)Lu, Mishra, Xia, Qiu, Chang, Zhu, Tafjord, Clark, and Kalyan]{lu2022learn}
Pan Lu, Swaroop Mishra, Tanglin Xia, Liang Qiu, Kai-Wei Chang, Song-Chun Zhu, Oyvind Tafjord, Peter Clark, and Ashwin Kalyan.
\newblock Learn to explain: Multimodal reasoning via thought chains for science question answering.
\newblock \emph{Advances in Neural Information Processing Systems}, 35:\penalty0 2507--2521, 2022.

\bibitem[Nallapati et~al.(2016)Nallapati, Zhou, Gulcehre, Xiang, et~al.]{nallapati2016abstractive}
Ramesh Nallapati, Bowen Zhou, Caglar Gulcehre, Bing Xiang, et~al.
\newblock Abstractive text summarization using sequence-to-sequence rnns and beyond.
\newblock \emph{arXiv preprint arXiv:1602.06023}, 2016.

\bibitem[Narayan et~al.(2018)Narayan, Cohen, and Lapata]{Narayan2018DontGM}
Shashi Narayan, Shay~B. Cohen, and Mirella Lapata.
\newblock Don't give me the details, just the summary! topic-aware convolutional neural networks for extreme summarization.
\newblock \emph{ArXiv}, abs/1808.08745, 2018.

\bibitem[Nijkamp et~al.(2022)Nijkamp, Pang, Hayashi, Tu, Wang, Zhou, Savarese, and Xiong]{nijkamp2022codegen}
Erik Nijkamp, Bo~Pang, Hiroaki Hayashi, Lifu Tu, Huan Wang, Yingbo Zhou, Silvio Savarese, and Caiming Xiong.
\newblock Codegen: An open large language model for code with multi-turn program synthesis.
\newblock \emph{arXiv preprint arXiv:2203.13474}, 2022.

\bibitem[Parrish et~al.(2021)Parrish, Chen, Nangia, Padmakumar, Phang, Thompson, Htut, and Bowman]{parrish2021bbq}
Alicia Parrish, Angelica Chen, Nikita Nangia, Vishakh Padmakumar, Jason Phang, Jana Thompson, Phu~Mon Htut, and Samuel~R Bowman.
\newblock Bbq: A hand-built bias benchmark for question answering.
\newblock \emph{arXiv preprint arXiv:2110.08193}, 2021.

\bibitem[Shetty and Adibi(2004)]{shetty2004enron}
Jitesh Shetty and Jafar Adibi.
\newblock The enron email dataset database schema and brief statistical report.
\newblock \emph{Information sciences institute technical report, University of Southern California}, 4\penalty0 (1):\penalty0 120--128, 2004.

\bibitem[Singh et~al.(2019)Singh, Natarajan, Shah, Jiang, Chen, Batra, Parikh, and Rohrbach]{singh2019towards}
Amanpreet Singh, Vivek Natarajan, Meet Shah, Yu~Jiang, Xinlei Chen, Dhruv Batra, Devi Parikh, and Marcus Rohrbach.
\newblock Towards vqa models that can read.
\newblock In \emph{Proceedings of the IEEE/CVF conference on computer vision and pattern recognition}, pages 8317--8326, 2019.

\bibitem[stability.ai(2024)]{zephyr_new}
stability.ai.
\newblock Zephyr.
\newblock \url{https://stability.ai/news/stablelm-zephyr-3b-stability-llm}, 2024.

\bibitem[Sun et~al.(2024)Sun, Huang, Wang, Wu, Zhang, Gao, Huang, Lyu, Zhang, Li, et~al.]{sun2024trustllm}
Lichao Sun, Yue Huang, Haoran Wang, Siyuan Wu, Qihui Zhang, Chujie Gao, Yixin Huang, Wenhan Lyu, Yixuan Zhang, Xiner Li, et~al.
\newblock Trustllm: Trustworthiness in large language models.
\newblock \emph{arXiv preprint arXiv:2401.05561}, 2024.

\bibitem[Team et~al.(2023)Team, Anil, Borgeaud, Wu, Alayrac, Yu, Soricut, Schalkwyk, Dai, Hauth, et~al.]{team2023gemini}
Gemini Team, Rohan Anil, Sebastian Borgeaud, Yonghui Wu, Jean-Baptiste Alayrac, Jiahui Yu, Radu Soricut, Johan Schalkwyk, Andrew~M Dai, Anja Hauth, et~al.
\newblock Gemini: a family of highly capable multimodal models.
\newblock \emph{arXiv preprint arXiv:2312.11805}, 2023.

\bibitem[Team et~al.(2024)Team, Mesnard, Hardin, Dadashi, Bhupatiraju, Pathak, Sifre, Rivi{\`e}re, Kale, Love, et~al.]{team2024gemma}
Gemma Team, Thomas Mesnard, Cassidy Hardin, Robert Dadashi, Surya Bhupatiraju, Shreya Pathak, Laurent Sifre, Morgane Rivi{\`e}re, Mihir~Sanjay Kale, Juliette Love, et~al.
\newblock Gemma: Open models based on gemini research and technology.
\newblock \emph{arXiv preprint arXiv:2403.08295}, 2024.

\bibitem[Tong et~al.(2024)Tong, Liu, Zhai, Ma, LeCun, and Xie]{tong2024eyes}
Shengbang Tong, Zhuang Liu, Yuexiang Zhai, Yi~Ma, Yann LeCun, and Saining Xie.
\newblock Eyes wide shut? exploring the visual shortcomings of multimodal llms.
\newblock \emph{arXiv preprint arXiv:2401.06209}, 2024.

\bibitem[Touvron et~al.(2023)Touvron, Martin, Stone, Albert, Almahairi, Babaei, Bashlykov, Batra, Bhargava, Bhosale, et~al.]{touvron2023llama}
Hugo Touvron, Louis Martin, Kevin Stone, Peter Albert, Amjad Almahairi, Yasmine Babaei, Nikolay Bashlykov, Soumya Batra, Prajjwal Bhargava, Shruti Bhosale, et~al.
\newblock Llama 2: Open foundation and fine-tuned chat models.
\newblock \emph{arXiv preprint arXiv:2307.09288}, 2023.

\bibitem[Wang et~al.(2024)Wang, Ma, Feng, Zhang, Yang, Zhang, Chen, Tang, Chen, Lin, et~al.]{wang2024survey}
Lei Wang, Chen Ma, Xueyang Feng, Zeyu Zhang, Hao Yang, Jingsen Zhang, Zhiyuan Chen, Jiakai Tang, Xu~Chen, Yankai Lin, et~al.
\newblock A survey on large language model based autonomous agents.
\newblock \emph{Frontiers of Computer Science}, 18\penalty0 (6):\penalty0 1--26, 2024.

\bibitem[Wang et~al.(2023)Wang, Li, Han, Nakov, and Baldwin]{wang2023not}
Yuxia Wang, Haonan Li, Xudong Han, Preslav Nakov, and Timothy Baldwin.
\newblock Do-not-answer: A dataset for evaluating safeguards in llms.
\newblock \emph{arXiv preprint arXiv:2308.13387}, 2023.

\bibitem[Xia et~al.(2024)Xia, Xing, Du, Yang, Feng, Xu, Yin, and Xiong]{xia2024fofo}
Congying Xia, Chen Xing, Jiangshu Du, Xinyi Yang, Yihao Feng, Ran Xu, Wenpeng Yin, and Caiming Xiong.
\newblock Fofo: A benchmark to evaluate llms' format-following capability.
\newblock \emph{arXiv preprint arXiv:2402.18667}, 2024.

\bibitem[Yang et~al.(2018)Yang, Qi, Zhang, Bengio, Cohen, Salakhutdinov, and Manning]{yang2018hotpotqa}
Zhilin Yang, Peng Qi, Saizheng Zhang, Yoshua Bengio, William~W Cohen, Ruslan Salakhutdinov, and Christopher~D Manning.
\newblock Hotpotqa: A dataset for diverse, explainable multi-hop question answering.
\newblock \emph{arXiv preprint arXiv:1809.09600}, 2018.

\bibitem[Yu et~al.(2023)Yu, Wang, Tu, Cao, Zhang-Li, Lv, Peng, Yao, Zhang, Li, et~al.]{yu2023kola}
Jifan Yu, Xiaozhi Wang, Shangqing Tu, Shulin Cao, Daniel Zhang-Li, Xin Lv, Hao Peng, Zijun Yao, Xiaohan Zhang, Hanming Li, et~al.
\newblock Kola: Carefully benchmarking world knowledge of large language models.
\newblock \emph{arXiv preprint arXiv:2306.09296}, 2023.

\bibitem[Zhang et~al.(2024{\natexlab{a}})Zhang, Xu, Zhao, Chen, Cao, and Yu]{zhang2024mobileenv}
Danyang Zhang, Hongshen Xu, Zihan Zhao, Lu~Chen, Ruisheng Cao, and Kai Yu.
\newblock Mobile-env: An evaluation platform and benchmark for llm-gui interaction, 2024{\natexlab{a}}.

\bibitem[Zhang et~al.(2024{\natexlab{b}})Zhang, Zeng, Wang, and Lu]{zhang2024tinyllama}
Peiyuan Zhang, Guangtao Zeng, Tianduo Wang, and Wei Lu.
\newblock Tinyllama: An open-source small language model.
\newblock \emph{arXiv preprint arXiv:2401.02385}, 2024{\natexlab{b}}.

\bibitem[Zhang et~al.(2023)Zhang, Lei, Wu, Sun, Huang, Long, Liu, Lei, Tang, and Huang]{zhang2023safetybench}
Zhexin Zhang, Leqi Lei, Lindong Wu, Rui Sun, Yongkang Huang, Chong Long, Xiao Liu, Xuanyu Lei, Jie Tang, and Minlie Huang.
\newblock Safetybench: Evaluating the safety of large language models with multiple choice questions.
\newblock \emph{arXiv preprint arXiv:2309.07045}, 2023.

\bibitem[Zheng et~al.(2023)Zheng, Chiang, Sheng, Zhuang, Wu, Zhuang, Lin, Li, Li, Xing, Zhang, Gonzalez, and Stoica]{zheng2023judging}
Lianmin Zheng, Wei-Lin Chiang, Ying Sheng, Siyuan Zhuang, Zhanghao Wu, Yonghao Zhuang, Zi~Lin, Zhuohan Li, Dacheng Li, Eric.~P Xing, Hao Zhang, Joseph~E. Gonzalez, and Ion Stoica.
\newblock Judging llm-as-a-judge with mt-bench and chatbot arena, 2023.

\bibitem[Zheng et~al.(2024)Zheng, Chiang, Sheng, Zhuang, Wu, Zhuang, Lin, Li, Li, Xing, et~al.]{zheng2024judging}
Lianmin Zheng, Wei-Lin Chiang, Ying Sheng, Siyuan Zhuang, Zhanghao Wu, Yonghao Zhuang, Zi~Lin, Zhuohan Li, Dacheng Li, Eric Xing, et~al.
\newblock Judging llm-as-a-judge with mt-bench and chatbot arena.
\newblock \emph{Advances in Neural Information Processing Systems}, 36, 2024.

\bibitem[Zhou et~al.(2023)Zhou, Lu, Mishra, Brahma, Basu, Luan, Zhou, and Hou]{zhou2023instruction}
Jeffrey Zhou, Tianjian Lu, Swaroop Mishra, Siddhartha Brahma, Sujoy Basu, Yi~Luan, Denny Zhou, and Le~Hou.
\newblock Instruction-following evaluation for large language models.
\newblock \emph{arXiv preprint arXiv:2311.07911}, 2023.

\end{thebibliography}

\newpage

\appendix

\section{Appendix}

\subsection{Dataset}

Datasets within MobileAIBench are summarized in Table~\ref{tab:dataset}. We select tasks with real-world Mobile use cases. These $20$ datasets cover NLP, Multi-modality and Trust\&Safety tasks, providing a comprehensive evaluation for mobile use cases. To support runable experiments on mobile devices, we down-sample the origional datasets to mostly $1000$ samples.

\begin{table}[h]
  \centering
  \caption{Summary of datasets integrated within MobileAIBench.}
  \scalebox{0.8}{
     \begin{tabular}{rrlrr}
     \toprule
    \multicolumn{1}{l}{\textbf{Task-type}} & \multicolumn{1}{l}{\textbf{Tasks}}  & \textbf{Datasets} & \multicolumn{1}{l}{\textbf{\# Samples}} & \multicolumn{1}{l}{\textbf{Avg. \# Length}}  \\
    \midrule
          \multicolumn{1}{l}{NLP}& \multicolumn{1}{l}{Question Answering} & Databricks-dolly \cite{DatabricksBlog2023DollyV2} &1000 &156.90  \\
          &       &        HotpotQA \cite{yang2018hotpotqa}   &1000 &443.59    
  \\
            \cdashline{2-5}
          & \multicolumn{1}{l}{Summarization}  & CNN/Daily Mail\cite{hermann2015teaching, nallapati2016abstractive}  &1000 &482.03  
 \\
          &       &        XSum \cite{Narayan2018DontGM}   &1000  &298.69       \\
          \cdashline{2-5}
           & \multicolumn{1}{l}{Text-to-SQL}  & Sql-create-context \cite{b-mc2_2023_sql-create-context} &1000   &18.61     \\
         % \cdashline{2-6}
          & \multicolumn{1}{l}{Language Understanding} & MMLU~\cite{hendrycks2020measuring} &1000 &94.82 \\
         % \cdashline{2-6}
          & \multicolumn{1}{l}{Math} & GSM8K~\cite{cobbe2021gsm8k} &1000 &72.35 \\
          % \cdashline{2-6}
          & \multicolumn{1}{l}{LLM Benchmarks} & Alpacaeval~\cite{alpaca_eval} &805 &28.56 \\
          & & MTBench~\cite{zheng2024judging} &80 &66.97 \\

     \midrule
    \multicolumn{1}{l}{Multi-modality} & \multicolumn{1}{l}{Direct Answer VQA} & VQA-v2~\cite{goyal2017making} & 1000 & 15.18  \\
        & & VizWiz~\cite{gurari2018vizwiz} & 1000 & 25.20 \\
        & & GQA~\cite{hudson2019gqa} & 1000 & 17.55 \\
        & & TextVQA~\cite{singh2019towards} & 1000 & 16.05 \\
        \cdashline{2-5}
        & \multicolumn{1}{l}{Multiple-Choice VQA} & ScienceQA~\cite{lu2022learn} & 1000 & 59.95 \\

    \midrule
    \multicolumn{1}{l}{Trust \& Safety} & \multicolumn{1}{l}{Truthfulness} & TruthfulQA~\cite{lin2021truthfulqa} &817 &71.23 \\
     & \multicolumn{1}{l}{Safety} & Do-Not-Answer~\cite{wang2023not} &1000 &14.33 \\
     & \multicolumn{1}{l}{Robustness} & Adversarial Instruction~\cite{sun2024trustllm} &600 &9.70 \\
     & \multicolumn{1}{l}{Fairness} & BBQ~\cite{parrish2021bbq} &1000 &67.35 \\
     & \multicolumn{1}{l}{Privacy} & Privacy Leakage~\cite{shetty2004enron} &150 &11.25  \\
     & \multicolumn{1}{l}{Ethics} & Social Chemistry 101~\cite{forbes2020social} &500 &22.05 \\
    \bottomrule
    \end{tabular}%
  }
  \label{tab:dataset}%
\end{table}%

\subsection{Experiment Model Selection}\label{sec:model_selection}

% \textbf{LLM Model Selection:} We select several LLMs of varied sizes to test the model performance and utilization on mobile devices. We category the selected models into large-size (>6B) and medium-size (2B-6B). For large-size LLMs, we select Llama2~\cite{touvron2023llama}, Mixtral-7B~\cite{jiang2023mistral} and Gemma-7B~\cite{team2024gemma} for comparison. Within medium-sized LLMs, we choose phi-2~\cite{javaheripi2023phi}, TinyLlama-1.1B~\cite{zhang2024tinyllama}, xgen2-3b, Gemma-2B~\cite{team2024gemma} and Stablelm-3B.

\textbf{LLM Model Selection:} We select several LLMs of varied sizes and architectures to evaluate their performance and utilization on mobile devices. The selected models are categorized into large-size (>6B) and medium-size (1B-6B) groups. For large-size LLMs, we include Llama2~\cite{touvron2023llama}, Mixtral-7B~\cite{jiang2023mistral}, and Gemma-7B~\cite{team2024gemma}. Among medium-sized LLMs, we choose Phi-2~\cite{javaheripi2023phi}, TinyLlama-1.1B~\cite{zhang2024tinyllama}, Gemma-2B~\cite{team2024gemma}, and StableLM-3B~\cite{zephyr_new}.

% \textbf{LMM Model Selection:} We categorize the selected models into three groups based on their parameter sizes, with two models in each category: 1) Large Models (around 7B parameters): Llava-v1.5-7B and BakLLava-7B; 2) Medium Models (around 3B parameters): Llava-phi-2 and Mobile-VLM-3B; 3) Relatively Small Models (around 2B parameters): Mobile-VLM-1.7B and Moondream2.

\textbf{LMM Model Selection:} We categorize the selected models into two groups: Large Models (> 6B parameters), including Llava-v1.5-7B and BakLLava-7B, and Medium Models (1B-6B parameters), which consist of Llava-phi-2, Mobile-VLM-3B, Mobile-VLM-1.7B, and Moondream2.

The LLMs considered for mobile testing include Phi-2, Gemma-2B, and TinyLlama-1.1B, all quantized to 4-bit to ensure compatibility with the iPhone 14's resource constraints. 

We note that serving LMMs on mobile devices is significantly more challenging than serving text-based LLMs due to their ensemble structure, larger model sizes, and complex inference processes. Therefore, we are not limiting the model selection to those fully supported by Llama.cpp, at this time. Instead, we select small-sized LMMs that are likely to be supported in the foreseeable future.
% For Efficiency/Utilization evaluation, we test the LLMs performance on iPhone14. LLMs for mobile testing (Phi2, Gemma, TinyLlama) are all quantized to 4-bit to satisfy the resource requirement for iPhone14.

\subsection{Evaluation Details}
Various evaluation metrics are considered for Standard NLP tasks, Multi-modality, and Trust \& Safety. The details of the same are provided below.
\subsubsection{Evaluation Prompt Templates}
\label{sec:prompt_template}
The evaluation prompts for various NLP datasets are listed in Table \ref{tab:evaluation-prompts-nlp}, while those for various multimodal datasets are listed in Table \ref{tab:evaluation-prompts-vqa}.
\subsubsection{Evaluation Metrics}
\label{sec:eval_metrics}
\begin{itemize}
    \item \textbf{Task: Question \& Answering}
    \begin{itemize}
        \item \textbf{Exact Match:} The Exact Match score evaluates the accuracy of a QA system by checking whether the predicted answer is exactly the same as the reference answer. This means that for each question, the predicted answer must match the ground truth answer exactly, including any formatting, punctuation, and whitespace.
        \item \textbf{F1 Score:} F1 score combines precision and recall into a single metric by taking their harmonic mean. We tokenize the ground truth and predictions, then we computer the precision and recall and finall compute the F1 score. 
    \end{itemize}

    \item \textbf{Task: Summarization}
    \begin{itemize}
        \item \textbf{Rouge-1:} The ROUGE-1 metric measures the overlap of unigrams (single words) between the candidate summary and the reference summary. 
        \item \textbf{Rouge-L:} The ROUGE-L metric evaluates the longest common subsequence (LCS) between the candidate summary and the reference summary. It measures the precision, recall, and F1 score based on the longest matching sequence of words, emphasizing the importance of order and continuity in the generated summaries.
    \end{itemize}

    \item \textbf{Task: Text-to-SQL}
    \begin{itemize}
        \item \textbf{SQL Parser:} The SQL query is converted into a graph where each node represents an SQL keyword, and its children represent the items associated with that keyword. This is implemented using a Python dictionary. To measure the SQL parser score, we evaluate the overlap between the ground truth graph and the predicted graph.
        \item \textbf{Levenshtein score:} The Levenshtein score, also known as the Levenshtein distance, measures the minimum number of single-character edits (insertions, deletions, or substitutions) required to change one word or string into another.
    \end{itemize}

    \item \textbf{Task: AlpacaEval}
    \begin{itemize}
        \item \textbf{Win-Rate:} Fraction of times predicted response is chosen over the predictions made by a baseline model. (In our case, baseline model is GPT-4)
    \end{itemize}

    \item \textbf{Task: MT-Bench}
    \begin{itemize}
        \item \textbf{Score:} Using GPT-4 in llm-as-a-judge framework, we ask the judge to score a given response on a scale of 10. (10 being the highest score)
    \end{itemize}

    \item \textbf{Task: Trust \& Safety}
    \begin{itemize}
        \item \textbf{Accuracy:} Using GPT-4o in llm-as-a-judge framework, we ask the judge to determine if the predicted answer is same as the ground truth answer.
    \end{itemize}
    \item \textbf{Task: VQA}
    \begin{itemize}
        \item \textbf{Score (single ground-truth):} For datasets with a single ground-truth answer per question (e.g., GQA, ScienceQA), we score each test sample based on exact match: $1$ if the prediction matches, otherwise $0$. Then, the score for each dataset is calculated by averaging the total test cases.
        \item \textbf{Score (multiple ground-truth):} For datasets with multiple human-provided ground-truth answers (e.g., VQA-v2, VisWiz, TextVQA), the accuracy score for each test case is calculated as: $\min\left(\frac{\text{number of matches}}{3}, 1\right)$. This follows the official evaluation design of VQA-v2.
    \end{itemize}
    
\end{itemize}

\begin{table}[h!]
\centering
\caption{Evaluation prompts for different NLP datasets.}

\begin{tabular}{ll}
\toprule
\textbf{Dataset} & \textbf{Evaluation Prompt} \\
\midrule
Llama 2 7B & \texttt{[INST] <<SYS>> \{system\} <</SYS>> \{prompt\} [/INST]} \\
Mistral 7B & \texttt{<s>[INST] \{system\} \{prompt\} [/INST]} \\
Gemma 7B & \texttt{<start\_of\_turn>user\textbackslash n\{system\} \{prompt\}<end\_of\_turn>\textbackslash n<start\_of\_turn>model\textbackslash n} \\
Zephyr 3B & \texttt{<|user|>\textbackslash n\{system\}\textbackslash n\{prompt\}<|endoftext|>\textbackslash n<|assistant|>\textbackslash n} \\
Phi2 3B & \texttt{\{system\} Instruct:\{prompt\}\textbackslash nOutput} \\
Gemma 2B & \texttt{<start\_of\_turn>user\textbackslash n\{system\} \{prompt\}<end\_of\_turn>\textbackslash n<start\_of\_turn>model\textbackslash n} \\
TinyLlama 1B & \texttt{<|system|>\textbackslash n\{system\}</s>\textbackslash n<|user|>\textbackslash n\{prompt\}</s>\textbackslash n<|assistant|>} \\
\bottomrule
\end{tabular}
\label{tab:evaluation-prompts-nlp}
\end{table}

% \begin{table}[h!]
% \centering
% \begin{tabular}{ll}
% \toprule
% \textbf{Dataset} & \textbf{Evaluation Prompt} \\
% \midrule
% Llama 2 7B & \texttt{[INST] <<SYS>> \{system\} <</SYS>> \{prompt\} [/INST]} \\
% Mistral 7B & \texttt{<s>[INST] \{system\} \{prompt\} [/INST]} \\
% Gemma 7B & \texttt{\<start\_of\_turn>user\\\{system\} \{prompt\}<end\_of\_turn>\\<start\_of\_turn>model\\} \\
% Zephyr 3B & \texttt{\\\{system\}\\\{prompt\}\\\\} \\
% Phi2 3B & \texttt{\{system\} Instruct:\{prompt\}\\Output} \\
% Gemma 2B & \texttt{<start\_of\_turn>user\\\{system\} \{prompt\}<end\_of\_turn>\\<start\_of\_turn>model\\} \\
% TinyLlama 1B & \texttt{\\\{system\}</s>\\\\\{prompt\}</s>\\} \\
% \bottomrule
% \end{tabular}
% \caption{Evaluation Prompts for Different NLP Datasets}
% \label{tab:evaluation-prompts-nlp}
% \end{table}

\begin{table}[h!]
\centering
\caption{Evaluation prompts for different VQA datasets.}

\begin{tabular}{ll}
\toprule
\textbf{Dataset} & \textbf{Evaluation Prompt} \\
\midrule
VQAV2 & \texttt{[Question]}\textbackslash n Answer the question using a single word or phrase. \\ 
VisWiz & \texttt{[Question]}\textbackslash n When the provided information is insufficient, respond with 'Unanswerable'. \\
& Answer the question using a single word or phrase. \\
GQA & \texttt{[Question]}\textbackslash n Answer the question using a single word or phrase. \\
TextVQA & \texttt{[Question]}\textbackslash n Answer the question using a single word or phrase. \\
ScienceQA & Context: \texttt{[context]}\textbackslash n Question:\texttt{[question]}\textbackslash n Options: (A)\texttt{[Option Content]} \\&(B) \texttt{[Option Content]}...\textbackslash n Answer with the option letter from the given choices directly. \\
\bottomrule
\end{tabular}
\label{tab:evaluation-prompts-vqa}
\end{table}

\subsection{Experiment Results}

\subsubsection{Utilization Experiments}\label{sec:utilization}
\begin{figure}[h]
\centering
    \includegraphics[width=0.24\textwidth]{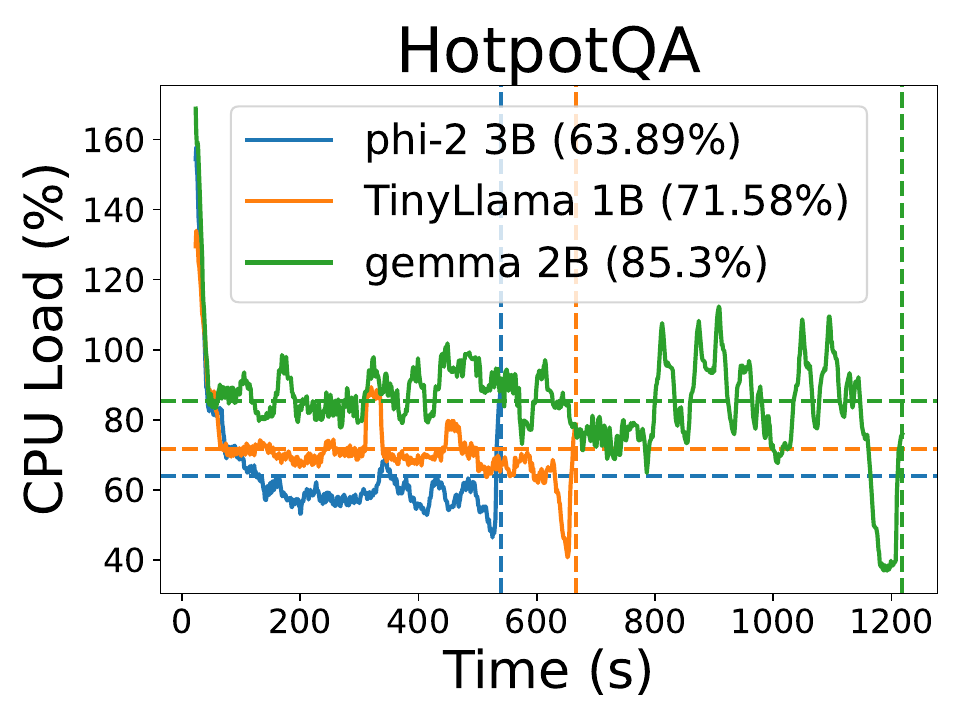}
    \includegraphics[width=0.24\textwidth]{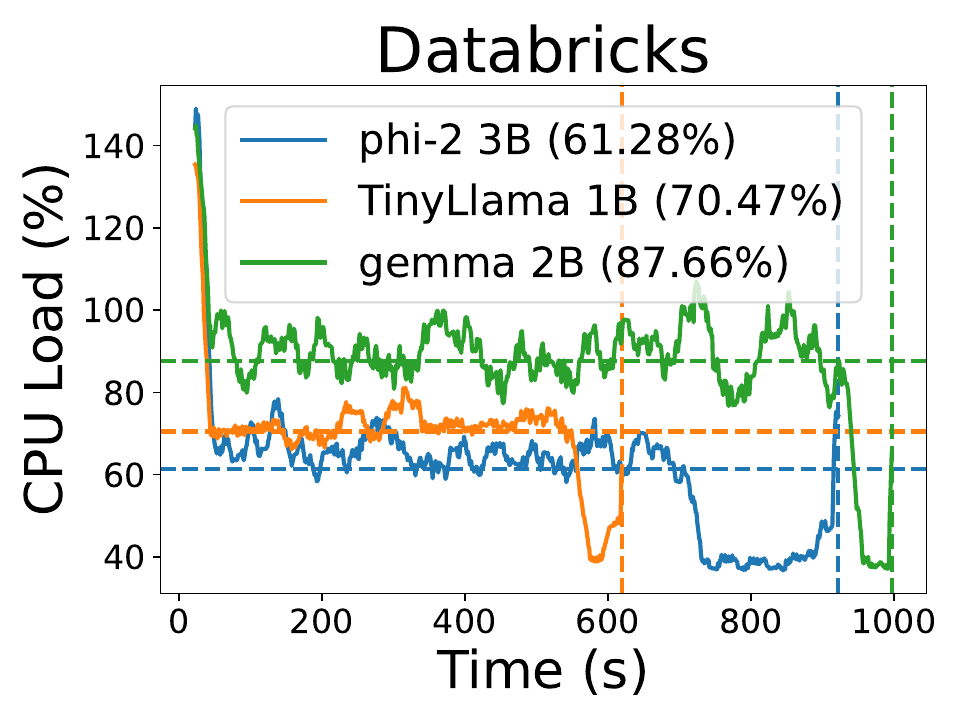}
    \includegraphics[width=0.24\textwidth]{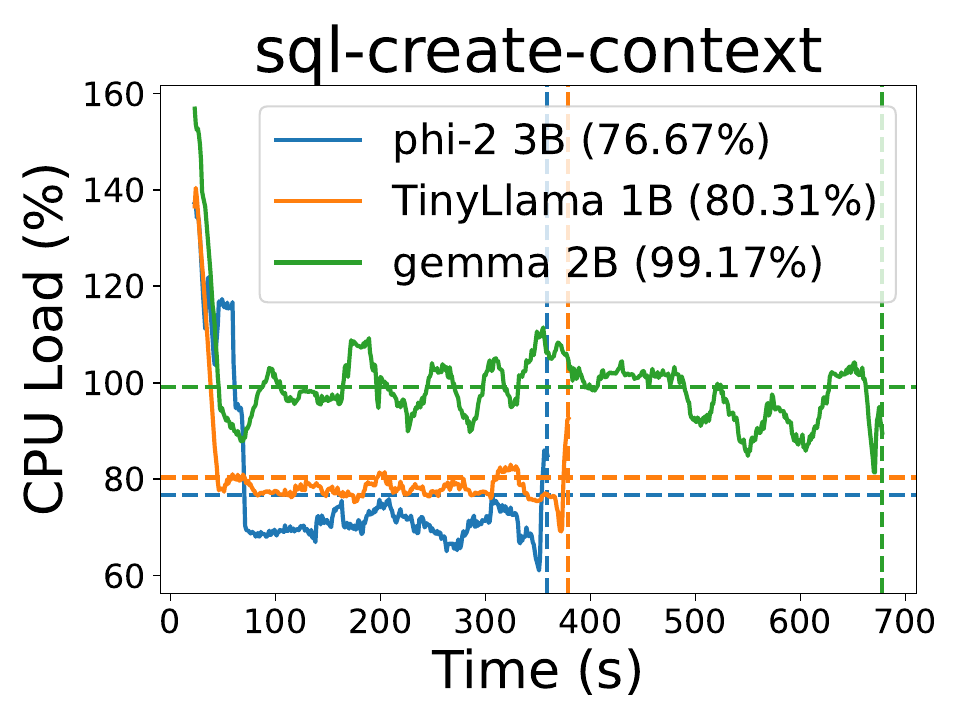}
    \includegraphics[width=0.24\textwidth]{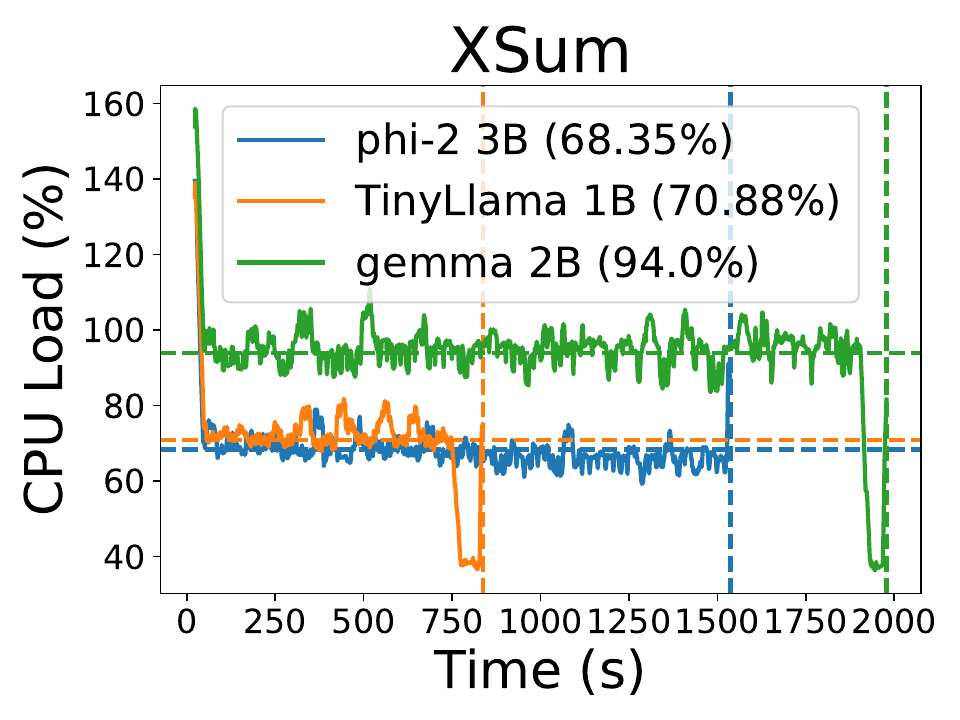}

    \includegraphics[width=0.24\textwidth]{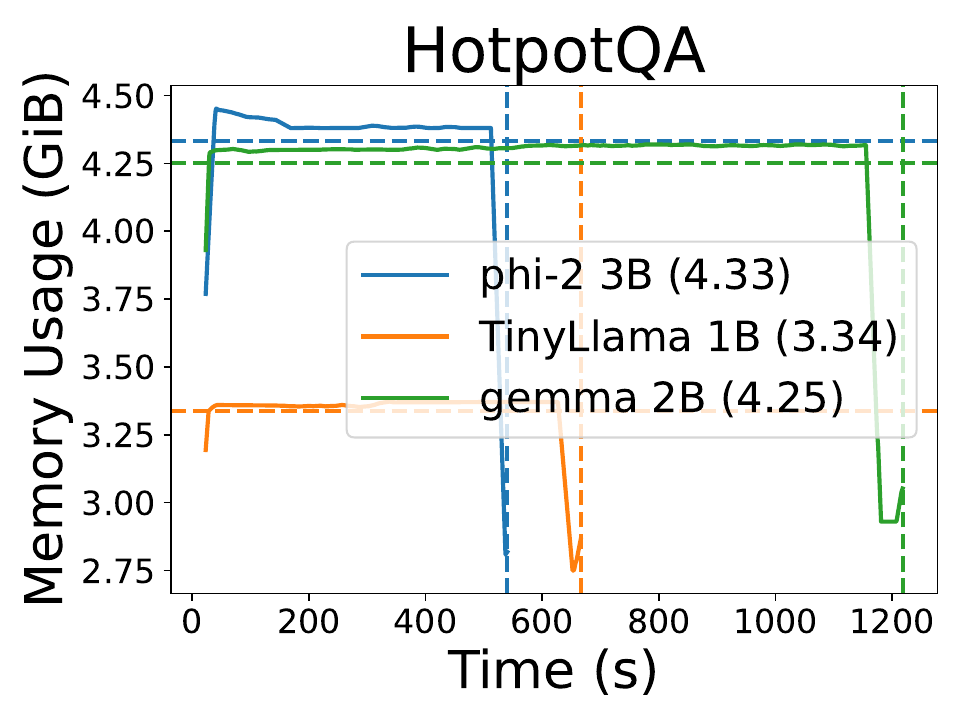}
    \includegraphics[width=0.24\textwidth]{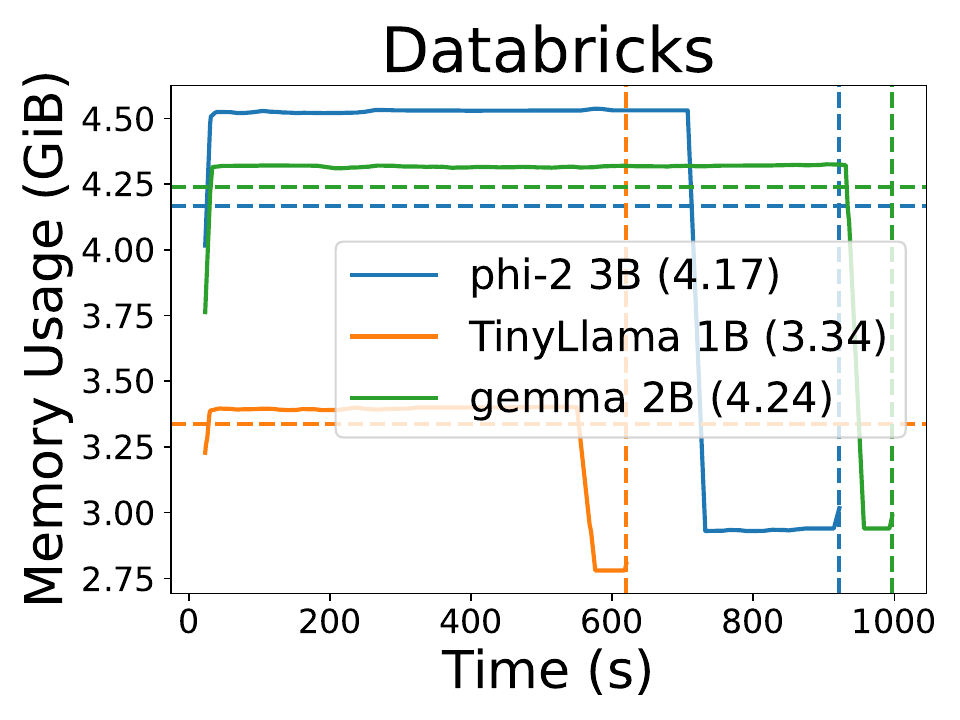}
    \includegraphics[width=0.24\textwidth]{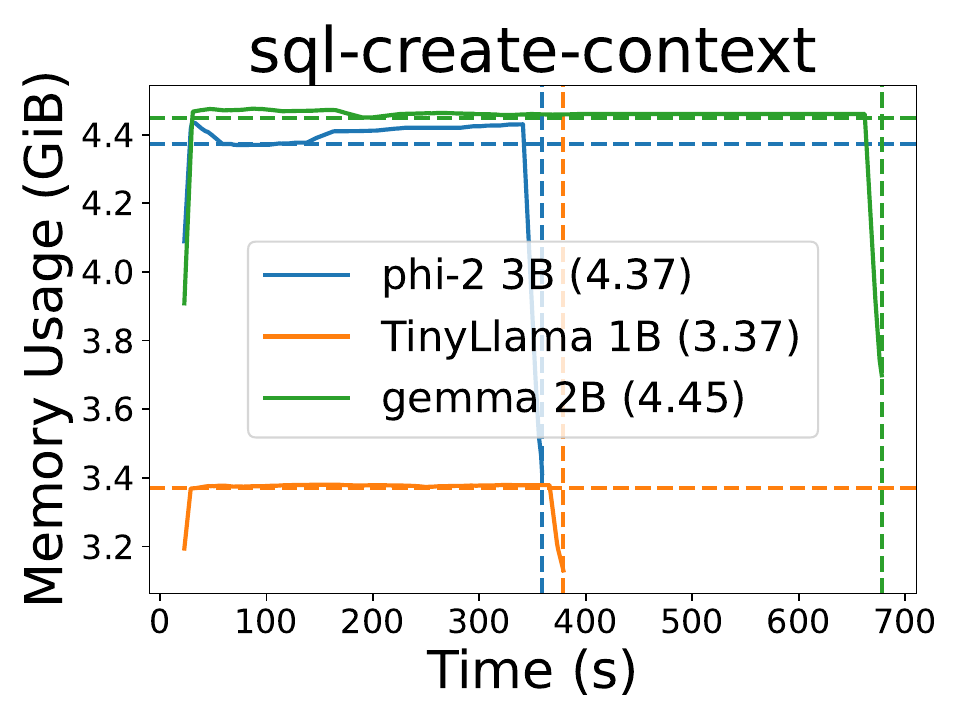}
    \includegraphics[width=0.24\textwidth]{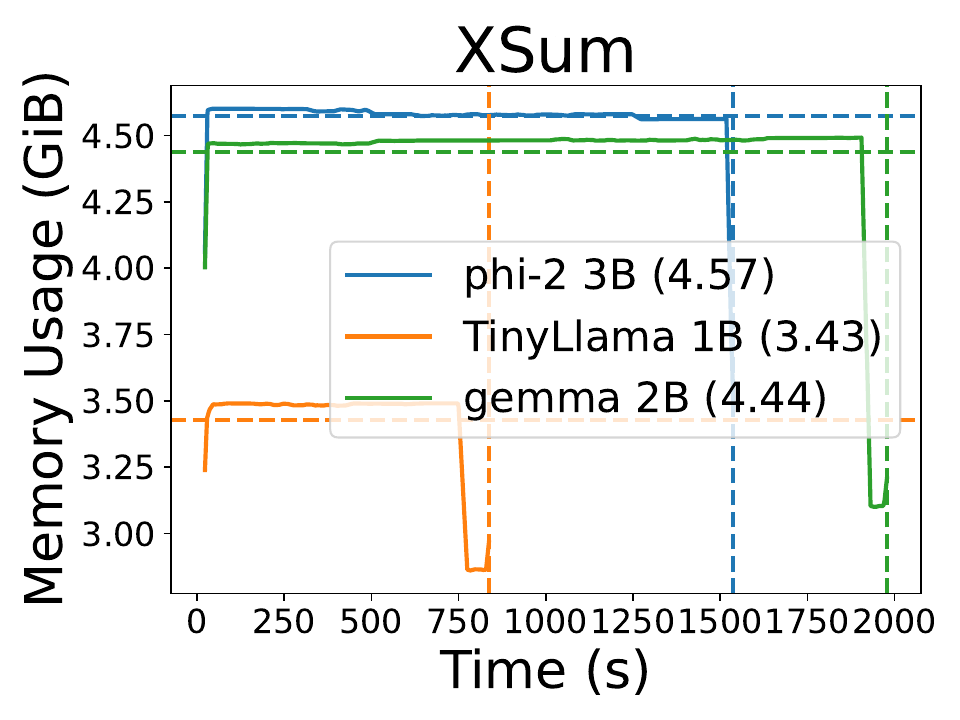}

\caption{CPU/Memory trace of different LLMs.}
\label{fig:trace}
\end{figure}

When running experiments on-device with MobileAIBench, we also obtain the trace files for CPU/Memory utilization. LLMs experiment trace files are shown in Figure~\ref{fig:trace}. From the utilization trace file, we can observe the changes when running LLMs/LMMs on device.
We can observe that on the four datasets, CPU utilization is not stable, and has a large fluctuations on the utilization curve. That is caused by the different CPU utilization when loading/inferencing samples.
However, the memory keeps nearly constant during inference, which shows models take the majority of memory and the memory cost for samples are relatively limited. Same trace files can also be obtained for LMM tasks, and we omit them for similar observations.

\subsubsection{Latency Analysis for LMMs on Intel CPU}
\label{sec:LMM_cpu}
Given the computational limitations of current mobile devices, it is challenging to conduct comprehensive on-device latency evaluations for all LMMs. To compare the latency of these models and provide information about their latency-related performance, we conduct latency evaluations on Intel CPUs of model type Intel(R) Xeon(R) CPU @ 2.20GHz. The results are shown in Table \ref{tab:LMM_latency}. 

The results indicate that quantizing the models to 8-bit and 4-bit levels generally improves inference speed. However, 3-bit quantization does not result in faster 
performance compared to the original model. 

Among all the models, smaller ones generally exhibit lower latency. Additionally, the results clearly demonstrate that the Mobile-VLM series has significantly lower latency compared to other models of similar size. According to the original paper \cite{chu2023mobilevlm}, the low latency of Mobile-VLM is attributed to the application of an additional convolutional layer after the visual tokens, which reduces the number of image tokens by a factor of four. This approach could be a viable strategy for developing latency-driven LMMs.
\begin{table}[]
\centering
\caption{Latency (time to first token) comparison of LMMs with different quantization levels.}
\begin{adjustbox}{width=0.55\linewidth}
\begin{tabular}{lcccc}
\toprule
Model            & 16bit & 8bit & 4bit & 3bit \\
\midrule
Llava-v1.5-7B    & 12.620 & 9.886 & 11.345 & 12.683 \\
BakLLaVA         & 14.099 & 10.515 & 11.986 & 13.851 \\
Llava-phi-2      & 6.513 & 5.398 & 6.025 & 7.030 \\
Mobile-VLM-3B    & \underline{2.917} & \underline{2.534} & \underline{2.690} & \underline{2.956} \\
Mobile-VLM-1.7B  & \textbf{2.003} & \textbf{1.889} & \textbf{1.901} & \textbf{2.054} \\
Moondream2       & 4.008 & 3.819 & 3.914 & 5.217 \\
\bottomrule
\end{tabular}
\end{adjustbox}
\label{tab:LMM_latency}
\end{table}

\end{document}